\def\paperID{3079}
\def\confName{ECCV}
\def\confYear{2024}

\def\paperTitle{Classification Matters: Improving Video Action Detection with Class-Specific Attention}

\def\abbvPaperTitle{Classification Matters: Improving VAD with Class-Specific Attention}

\def\suppPaperTitle{Classification Matters: Improving VAD with Class-Specific Attention \\
\normalfont \textit{---Supplementary Material---}}

\def\authorBlock{
    Jinsung Lee\inst{1 \thanks{Work done while doing an internship at NAVER Cloud.}}\orcidlink{0009-0007-0782-8651} \and
    Taeoh Kim \inst{2}\orcidlink{0000-0001-7252-5525} \and
    Inwoong Lee \inst{2}\orcidlink{0000-0003-4356-7616} \and
    Minho Shim \inst{2}\orcidlink{0000-0002-9637-4909} \and \\
    Dongyoon Wee \inst{2}\orcidlink{0000-0003-0359-146X} \and
    Minsu Cho \inst{1}\orcidlink{0000-0001-7030-1958} \and
    Suha Kwak \inst{1}\orcidlink{0000-0002-4567-9091}
}

\def\authorInstitutes{
    Pohang University of Science and Technology (POSTECH), South Korea\\ \and
    NAVER Cloud, South Korea \\
    \url{https://jinsingsangsung.github.io/ClassificationMatters/}
}

\def\abbvAuthors{J.~Lee et al.}

\newif\ifreview 
\newif\ifcamera \newcommand{\cameraready}{\cameratrue}
\newif\ifmobile 
\cameraready
\documentclass[runningheads]{llncs}
\ifreview \usepackage[review,year=\confYear,ID=\paperID]{eccv} \fi
\ifcamera \usepackage{eccv} \fi
\ifmobile \usepackage[mobile]{eccv} \fi

\usepackage{eccvabbrv}

\usepackage{graphicx}
\usepackage{amsmath}	
\usepackage{amssymb}	
\usepackage{booktabs}
\usepackage{pifont}
\usepackage{microtype}
\usepackage{epsfig}
\usepackage{caption}
\usepackage{float}
\usepackage{placeins}
\usepackage{color, colortbl}
\usepackage{stfloats}
\usepackage{enumitem}
\usepackage{tabularx}
\usepackage{bbm}
\usepackage{xstring}
\usepackage{multirow}
\usepackage{xspace}
\usepackage{url}

\usepackage[subrefformat=simple,labelformat=simple,font=scriptsize]{subcaption}

\usepackage[hang,flushmargin]{footmisc}
\usepackage[accsupp]{axessibility}  

\newcommand{\cmark}{\ding{51}}
\newcommand{\xmark}{\ding{55}}
\newcommand{\hhh}{\rule[.2ex]{1em}{0.1pt}}
\DeclareMathOperator*{\argmin}{argmin}  

\usepackage[pagebackref,breaklinks,hidelinks, colorlinks, citecolor=eccvblue]{hyperref}

\usepackage{orcidlink}

\let\oldmaketitle\maketitle
\renewcommand{\maketitle}{\oldmaketitle\setcounter{footnote}{0}}

\begin{document}

\title{\paperTitle}

\titlerunning{\abbvPaperTitle}

\author{\authorBlock}
\authorrunning{\abbvAuthors}
\institute{\authorInstitutes}

\maketitle

\begin{abstract}
    Video action detection (VAD) aims to detect actors and classify their actions in a video.
    We figure that VAD suffers more from classification rather than localization of actors.
    Hence, we analyze how prevailing methods form features for classification and find that they prioritize actor regions, yet often overlooking the essential contextual information necessary for accurate classification.
    Accordingly, we propose to reduce the bias toward actor and encourage paying attention to the context that is relevant to each action class.
    By assigning a class-dedicated query to each action class, our model can dynamically determine where to focus for effective classification.
    The proposed model demonstrates superior performance on three challenging benchmarks with significantly fewer parameters and less computation.
  
\end{abstract}
\section{Introduction}
\label{sec:intro}

Video action detection (VAD) is the task of identifying actors and categorizing their activities in a video.
It has recently attracted increasing attention due to its broad range of applications, such as surveillance video analysis or sports activity recognition.
Since a video is a sequence of images, it is not surprising that solutions to video understanding tasks, including VAD, have been developed largely based on image recognition models.
In particular, since VAD resembles object detection (OD) in an image, a large number of existing VAD models consider the task as an extension of OD and follow common OD pipelines accordingly~\cite{peng2016multi, feichtenhofer2020x3d, wu2020context, wu2019long, pan2021actor, saha2016deep,sun2018actor, chen2021watch, kopuklu2019you, sui2023simple,zhao2022tuber}.

\begin{figure}[t!]
     \centering\
     \includegraphics[width=0.70\textwidth]{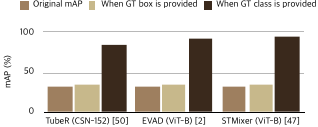}
     \caption{Detection performance changes of the state-of-the-art methods (i.e., TubeR~\cite{zhao2022tuber}, EVAD~\cite{chen2023efficient}, STMixer~\cite{wu2023stmixer}) on AVA~\cite{gu2018ava} when ground-truth boxes or class labels are given.
     }
     \label{Classification is important}
\end{figure}

Unfortunately, such straightforward extensions of OD are often not optimal for VAD due to the distinct nature of VAD from OD: \emph{all instances conducting actions in VAD are humans}.
Action localization in VAD focuses on identifying human-shaped objects only, making it considerably simpler than localizing arbitrary objects in OD~\cite{gu2018ava}.
In contrast, the classification of actions in VAD is significantly more difficult than that of OD.
Unlike object classification in OD, which usually depends more on general appearances of objects, action classification in VAD requires identifying fine-grained details in both appearance and motion since different action classes are all conducted by humans, and thus, their general appearances are often not distinguishable clearly.
We empirically verify the crucial role of classification in improving VAD performance.
As shown in Fig.~\ref{Classification is important}, for all the three latest VAD models we tested, providing ground-truth (GT) class labels consistently yields more substantial improvement than providing GT bounding boxes.
This result implies that the room for improvement in VAD is mainly occupied by classification rather than localization.
However, only a few have paid attention to such inherent challenge in classification for VAD~\cite{gu2018ava, chen2021watch, sun2018actor}.

\setlength{\textfloatsep}{11pt plus 0pt minus 5pt}
In this work, we propose a new model architecture that aims to improve the classification performance for VAD. 
Our model first localizes each actor by attending features globally and then seeks local regions that are informative for identifying its action class.
This procedure enables our model to actively focus on local regions that provide greater assistance in classification, such as fine details (\eg, a cigarette for the \textsl{`smoke'} action class) or other people interacting with the actor (\eg, a speaker for the \textsl{`listen to'} action class).
To this end, we introduce \emph{class queries}, each of which separately holds essential information about each action class.
These class queries learn to identify if a specific action class takes place within the scene.
Specifically, we first construct a feature map that encompasses the interaction between each actor and the global context.
Then, each class query is learned to be highly similar with features on a particular region of the feature map relevant to the action class, so that it extracts rich and fine-grained features that are dedicated to each actor and each class.
These features help the model identify details at a granular level and context necessary for action classification, while also offering interpretable attention maps for each class that support the model's decision-making concurrently.
The consequent attention map illustrates the model's ability to capture details and interactions relevant to the action happening in the scene, and the model seeks over regions that are not bounded by the actor box.
We additionally introduce components that notably enhance the utilization of class queries, aiding in capturing details and ensuring specificity to individual actors.

Our model is evaluated on three conventional benchmarks in the field: AVA~\cite{gu2018ava}, JHMDB51-21~\cite{jhuang2013towards}, and UCF101-24~\cite{soomro2012ucf101}.
Among the models with comparable backbones, our model shows superior performance while being more efficient.
Current best performing models~\cite{chen2023efficient, wu2023stmixer} predict an instance per frame given a clip of multiple frames, and thus adopt the sliding window strategy to localize an actor in a whole video.
In contrast, our model constructs the entire spatio-temporal tube of an actor through a single feed-forward pass, which better aligns with the objective of VAD and also leads to higher computational efficiency, in particular when dealing with longer video clips.

In this paper, we present four key contributions.
\begin{itemize}
    \item We analyze how existing methods in VAD process features for classification, and perform a detailed investigation into their problematic behaviors.
    \item We introduce a novel classification module, \emph{Classifying Decoder Layer}, that effectively combines context, actor, and class queries to construct classification features for each action class.
    \item We provide additional components, \emph{3D Deformable Transformer Encoder} and \emph{Localizing Decoder Layer}, that augment our classification module and significantly boost the model's performance.
    \item Our model outperforms existing methods on challenging benchmarks with greater efficiency.
\end{itemize}
\section{Related work}
\label{sec:related}

\textbf{Video action detection.}
Video action detection involves analyzing a video clip and generating actor-specific spatio-temporal tubes while simultaneously predicting the actions being performed by the actors.
Earlier studies~\cite{weinzaepfel2015learning, peng2016multi, kopuklu2019you} try to combine spatial and temporal information to obtain a good actor representation and pass it directly to box regression and classification layers.
As advancements in object detection techniques continue, it has become common to employ 2D person detectors to extract actors from the scene and utilize these features for classification~\cite{feichtenhofer2019slowfast, feichtenhofer2020x3d, tong2022videomae, wang2023videomae}.
On the other side, a few studies~\cite{zhao2022tuber} attempt to capture spatio-temporal tubes instead of detecting actors frame-by-frame.
Our approach adopts the tube-based process not only to embrace the temporal property of the task but also to enhance the computational efficiency.

\noindent
\textbf{Transformer-based architectures.}
In the field of object detection, DETR~\cite{carion2020end} has suggested a framework that elevates cross-attention between queries and image features to capture both object location and its class effectively.
With its notable performance and convenience, the advent of DETR has sparked the emergence of various adaptations~\cite{zhu2020deformable, meng2021conditional, wang2022anchor, li2022dn}.
Among all, DAB-DETR~\cite{liu2022dab} is the most relevant research to our work, as it introduces a modulating function that adjusts the positional information of queries.
This approach enhances the cross-attention mechanism, allowing for a more comprehensive representation that encompasses the width and height of the anchor box prior.
Our method utilizes this positional prior differently by offering class queries the clues for specifying actor, providing subtle guidance about which instance a class query should be referring to.

\noindent
\textbf{Methods that tackle the classification problem.}
To achieve better action classification score, extensive research has been conducted on methods that consistently explore the relationship between the actor and the contextual information.
In previous studies, this context has been categorized into two types: actor-actor relationship and actor-context relationship, and the methods used to establish these relationships often exhibit slight variations.
In order to obtain the relationship between the two, either detected instances and the global features are concatenated~\cite{wu2020context, pan2021actor, sun2018actor} or fed to a transformer module~\cite{tang2020asynchronous, zhao2022tuber, josmy2022holistic, chen2023efficient, herzig2022object}.
While the transformer module is an advanced design to take instances' relations into account, previous methods based on transformers suffer from the problem where the classification process attends to the regions that are prone to becoming biased toward near the actor regions.
In the subsequent section, we delve into a detailed discussion of this issue describing its reason and implications.
\section{Background} \label{problem_of_cross-attention}

\begin{figure}[t!]
     \centering
     \includegraphics[width=\textwidth]{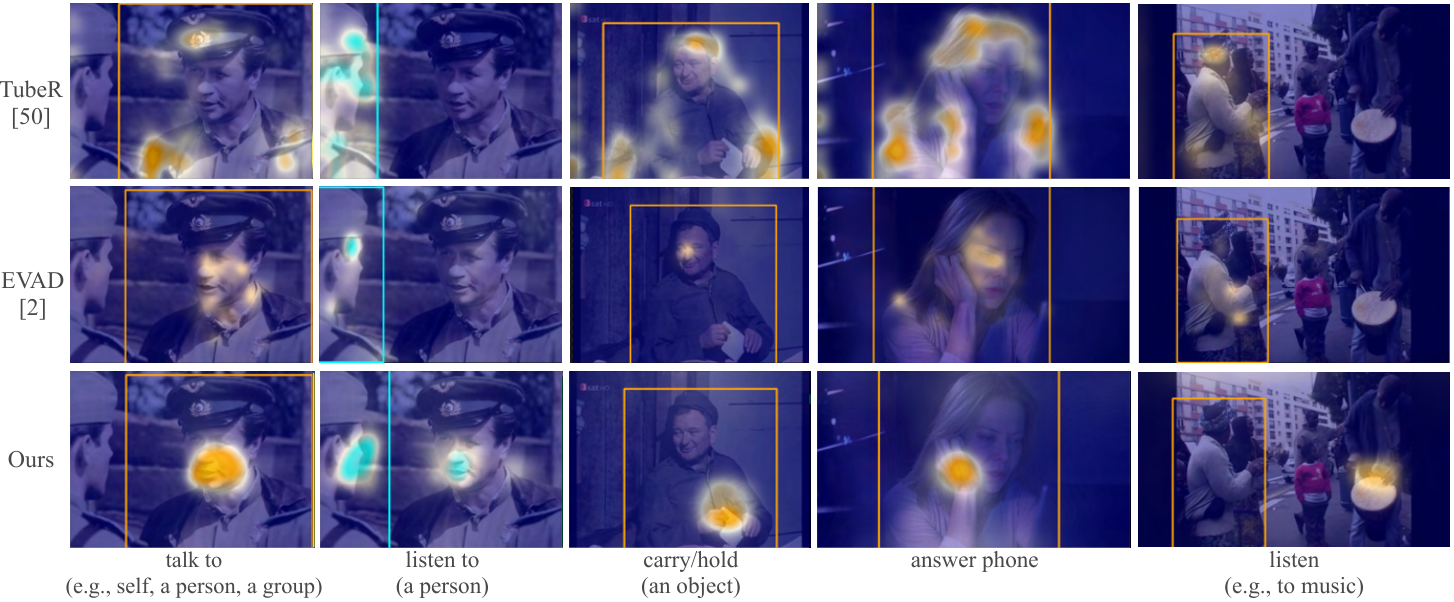}
     \caption{
     Sample detection results and classification attention maps of the previous transformer-based model, TubeR~\cite{zhao2022tuber}, EVAD~\cite{chen2023efficient}, and our model.
     Each attention map signifies the regions where the model attends to classify the action of the actor marked in the bounding box of the same color.
     Since our model creates an attention map for each class, we mark the corresponding label under the map. Best viewed in color.}
     \label{location_biased_detections}
\end{figure}

Following the popular OD models~\cite{carion2020end, zhu2020deformable, li2022dn}, it has become a de-facto standard in VAD to employ the transformer architecture~\cite{vaswani2017attention} in which query that represents an actor (referred to as the actor feature in VAD) gathers information from the video feature map (referred to as the context feature in VAD).
In particular, recent VAD models with this architecture~\cite{zhao2022tuber, chen2023efficient} utilize transformers to search regions that are relevant to each actor, and use the resulting attention map to create their classification features.

However, this structure makes the classification of prevailing methods become biased towards the context features near actor's location.
Since their output classification features are derived from a single attention map, all action classes share the same information of the context feature.
Thus, instead of understanding class-specific knowledge, the transformer weights are often trained only to embed commonly shared semantics across different classes.
As mentioned in Sec.~\ref{sec:intro}, action classes of VAD share an obvious semantic element, \emph{the actor}, and this characteristic inevitably forces their models to include more information related to actor itself, leading to higher attention values near actor regions.
Such classification feature may limit the model's observation scope to actor regions, which results in the model missing important regions that aid in classification.

The attention maps of TubeR~\cite{zhao2022tuber} depicted in Fig.~\ref{location_biased_detections} clearly illustrate the identified issue:
the attention is primarily concentrated on the actor's boundary regardless of the action the actor performs.
Similarly, in the case of EVAD~\cite{chen2023efficient}, the attention is mainly distributed over actor's face and body parts, yet critical regions essential for classification are overlooked.
Furthermore, both prior methods fail to extend their attention beyond the actor's bounding box, where crucial contexts that provide clues to distinguish action classes are located.
The activated regions may indicate the commonly shared semantics across different classes, but their lack of class-specificity leads to the model missing important clues for classification even when such clues are on the actor's body.
Therefore, we aim to increase the class specificity of the classification feature and improve the classification performance.

\section{Proposed method}
\label{sec:method}

\begin{figure*}[t!]
     \centering
     \includegraphics[width=0.98\textwidth]{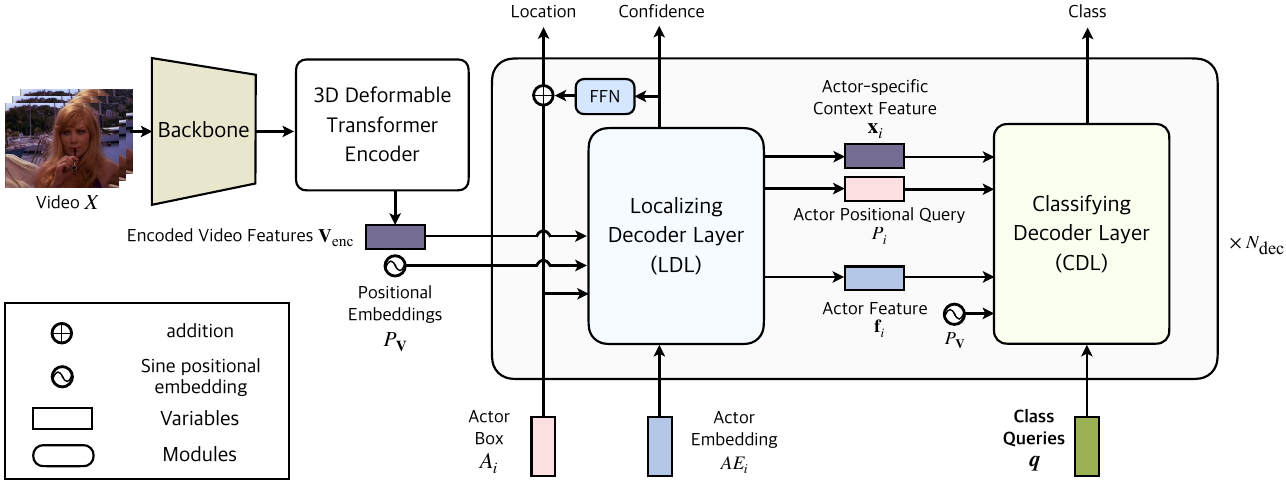}
      \caption{Overview of the proposed model}
     \label{figure3}
\end{figure*}

In this section, we introduce our model that addresses the aforementioned challenges.
Fig.~\ref{figure3} provides an overview of the model's architecture, consisting mainly of a backbone, transformer encoder, and transformer decoder.
Given an input video clip $X \in \mathbb{R}^{T \times H_0 \times W_0 \times 3}$ of RGB frames and $N_c$ action classes, our model operates as a function that takes $X$ as input and outputs $\hat{Y} = \bigl\{ \bigl(\hat{\mathbf{B}}_i, \hat{\mathbf{C}}_i \bigr) | \hat{\mathbf{B}}_i \in \mathbb{R}^{T \times 4}, \hat{\mathbf{C}}_i \in [0, 1]^{T \times N_c}, \ i \in [1, N_X] \bigr\}$, where $(T, H_0, W_0)$ is the temporal length, height and width of the input video clip, $\bigl(\hat{\mathbf{B}}_i, \hat{\mathbf{C}}_i \bigr)$ represents a spatio-temporal tube and per-frame action class prediction for the $i$-th actor, and $N_X$ is the total number of actors appearing in the video $X$.

The key component of the model lies in the classification module, which we denote as \emph{Classifying Decoder Layer} (CDL).
It distinguishes itself from other approaches through the utilization of \emph{class queries}, which are learnable embeddings designed to encapsulate information specific to each class label.
Class queries help the model's classification in two aspects.
First, they allow transformer architecture to create more variation between features that represent different action classes and mitigates the issue of being biased towards common semantics (\ie, the actor) of multiple classes.
Thus, class queries provide opportunities to explore beyond actor locations (\eg, \textsl{`listen to (a person)'} and \textsl{`listen (e.g., to music)'} in the third row of Fig.~\ref{location_biased_detections}).
Second, they acknowledge diverse characteristics of each class and grant the model more opportunities to browse over regions that are particularly conditioned to individual classes.
More details into the use of class queries are presented in Sec.~\ref{sec:decoder}.

CDL is strategically placed within the transformer decoder layers to enable access to features crucial for classification while allowing the acquisition of enriched information as layers are stacked.
To ensure that CDL receives informative features, we introduce essential modifications to the conventional transformer encoder and decoder layers, hereby referred to as \emph{3D Deformable Transformer Encoder} and \emph{Localizing Decoder Layer} (LDL), respectively.
In the following sections, we will delve into how each module operates, providing insights into their functions and contributions to the overall framework.

\subsection{Backbone and transformer encoder}
The input $X$ is passed through the backbone network and 1D convolutions, and we integrate the outputs from intermediate layers to produce multi-scale feature maps $\mathbf{V} = \bigl\{\boldsymbol{v}^l \in \mathbb{R}^{T_l \times H_l \times W_l \times D} | \ l\in [1, L] \bigr\}$, where $L$ is the number of feature map scales and $D$ is the output channel dimension.
Given the established benefits of utilizing multi-scale feature maps for effectively capturing various levels of semantics and details~\cite{lin2017_fpn, wang2017stagewise, pang2020multi, guo2020augfpn}, we leverage the multi-scale feature maps to achieve more precise classification.
Accordingly, we modify the ordinary transformer encoder to efficiently handle these feature maps,
which we dubbed \emph{3D Deformable Transformer Encoder}.
It processes the feature maps $\mathbf{V}$ and produces encoded feature maps $\mathbf{V}_{\textnormal{enc}}^{'}$ of the same shapes.
To mitigate the computational memory demands associated with multi-scale spatio-temporal features, we draw inspiration from Deformable DETR~\cite{zhu2020deformable}, where a query gets encoded with the features gathered from distant points that are determined by offsets and weights generated by the query itself.
Since the original work utilizes $(\Delta{h}, \Delta{w})$-shaped 2-dimensional offsets for its encoder, we extend this offset to $(\Delta{t}, \Delta{h}, \Delta{w})$ so that the query can be encoded with temporally distant features.
A detailed logic of the encoder is described in Sec.~C of the supplementary materials.

After $\mathbf{V}$ go through the encoder layers, the output feature maps $\mathbf{V}_{\textnormal{enc}}^{'}$ are scaled by interpolation to match the same spatio-temporal dimension across different levels:
$\mathbf{V}_{\textnormal{enc}} = \bigl\{\boldsymbol{v}_{\textnormal{enc}}^l \in \mathbb{R}^{T \times H \times W \times D} | \ l\in [1, L] \bigr\}$.
Note that the temporal dimension is recovered to the original length $T$ in order to output boxes and class labels for each timestep.

\begin{figure}[t!]
    \begin{subfigure}{0.5\textwidth}
        \centering
        \includegraphics[width=0.9\linewidth]{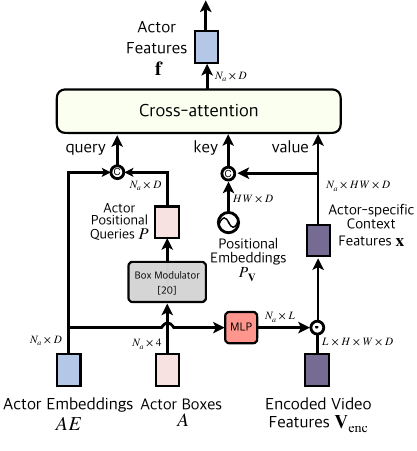} 
        \caption{Localizing Decoder Layer (LDL)}
        \label{fig:LDL_simple}
    \end{subfigure}
    \begin{subfigure}{0.49\textwidth}
        \centering
        \includegraphics[width=0.85\linewidth]{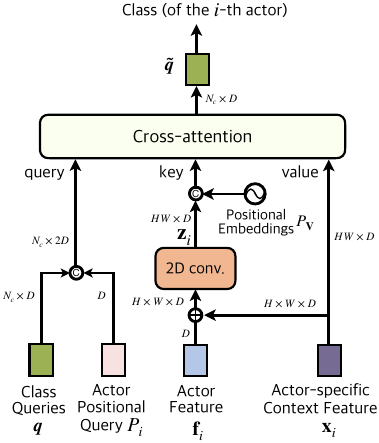}
        \caption{Classifying Decoder Layer (CDL)}
        \label{fig:CDL_simple}
    \end{subfigure}

\caption{Structure of transformer decoder layers of our model. We use 
{\scriptsize \textcircled{${\texttt{\text{\scriptsize {c}}}}$}}, $\odot$, and $\oplus$ to indicate concatenation, multiplication, and summation. In \subref{fig:CDL_simple}, we denote variables with the actor index $i$ to describe the process simpler.}
\label{fig:Decoder_layers}
\end{figure}

\subsection{Transformer decoder}\label{sec:decoder}
The decoder architecture serves as the fundamental design that embodies our objective of achieving improved classification.
It is divided into two modules: localizing decoder layer (LDL) and classifying decoder layer (CDL).
LDL specializes in gathering actor-related features from the encoded feature maps $\mathbf{V}_{\textnormal{enc}}$ and constructs localization features.
In contrast, CDL leverages the intermediate outputs of LDL, alongside class queries, to generate classification features.
The subsequent sections elaborate on each module.
Note that we simplify the notation by omitting the time index $t$ and focus on how an individual frame is processed within the decoder layer.
\subsubsection{Localizing decoder layer (LDL).}
LDL aims to construct features containing the information related to actors and provides informative features to CDL.
In a nutshell, LDL resembles the decoder layer of DETR~\cite{carion2020end}: it consists of a cross-attention layer that takes learnable queries and the encoder output $\mathbf{V}_{\textnormal{enc}}$ as its inputs to embed the actor information.
Though, it differs from the original DETR decoder in two aspects.
First, it constructs query and key by concatenating the \textit{content} and \textit{spatial} parts~\cite{meng2021conditional}, where each part plays a different role in embedding actor's appearance feature and actor's positional feature, respectively.
Such a design is essential in our model since class queries utilize actor-specific positional features for accurate classification, which we will describe in the next section.
Second, it aggregates multi-scale feature maps to create feature maps that are specific to each actor.
Due to the space limit, we briefly explain the roles and implications of LDL's input and output, and encourage readers to refer to Sec.~D of the supplementary material for details.

We describe the structure of LDL in Fig.~\ref{fig:LDL_simple}.
Let $N_a$ denote the number of actor candidates, then LDL takes as its query input $A \in \mathbb{R}^{N_a \times 4}$ and $AE \in \mathbb{R}^{N_a \times D}$, which we each denote as \textit{actor box} and \textit{actor embedding}.
$A$ and $AE$ are utilized as the spatial part and the content part of the input query, and learn to embed appearance and positional information of actors, respectively.
Thus, the output of LDL, which we denote as \textit{actor features} $\mathbf{f} \in \mathbb{R}^{N_a \times D}$, results in containing the information related to each actor.

During its process, we transform $A$ from coordinate space to $D$-dimensional space so that we can construct the \textit{actor positional queries} $P$, which act as the spatial part of the input query while containing spatial information of the actor.
At the same time, the multi-scale feature map $\mathbf{V}_{\textnormal{enc}}$ is aggregated to obtain single-scale feature map for each actor.
The weight utilized for the aggregation is acquired by applying a linear layer to $AE$, adding actor-specificity to key and value.
We denote the aggregated output as \textit{actor-specific context feature} $\mathbf{x}$.

\subsubsection{Classifying decoder layer (CDL).} \label{sec:Classfier}
CDL is designed to enable the model to selectively attend to class-specific information.
As mentioned in Sec.~\ref{problem_of_cross-attention}, the process for constructing classification features in the contemporary approaches~\cite{chen2023efficient, zhao2022tuber} introduces bias towards the actor's body parts.
By introducing class queries, our decoder determines on which context the model should focus for each action class, thus allowing for an expanded scope of observation that may extend beyond the bounding box of the actor.
However, adopting class queries in a naïve manner leads to the problem of \emph{actor-agnostic activation}; class queries may activate the class-specific context belonging to wrong actors, thereby gathering clues that are not relevant to the target actor the model aims to classify (Fig.~\ref{wopq}(a)).
Therefore, the objective of CDL is to enhance specificity of both actor and class simultaneously.
As class queries improve the class-specificity, we additionally incorporate actor positional queries $P$ and actor-specific context features $\mathbf{x}$ to address the actor-specificity.

We present the structure of CDL in Fig.~\ref{fig:CDL_simple}.
As CDL creates classification features for each actor in a parallel manner, we index features that correspond to each actor with $i \in [1, N_a]$ and describe how the module operates for the $i$-th actor.
CDL is a cross attention layer whose input query is comprised with class queries $\boldsymbol{q} \in \mathbb{R}^{N_c \times D}$ and the actor positional query $P_i \in \mathbb{R}^{D}$.
Since the input query is a combination of both class-specific and actor-specific features, the output feature of CDL grants information dedicated to each class and the actor concurrently.
To further enhance the actor-specificity of the output, we utilize the actor-specific context feature $\mathbf{x}_i$ as the input key and value of CDL.
We first broadcast the actor feature $\mathbf{f}_i \in \mathbb{R}^D$ to match the spatial dimension of $\mathbf{x}_i \in \mathbb{R}^{H \times W \times D}$, and take sum of these two features to pass it through subsequent convolutional layers.
Hence, the resulting feature map, which we denote as $\mathbf{z}_i \in \mathbb{R}^{HW \times D} $, represents the interaction that happens between the $i$-th actor and the context.
Finally, $\mathbf{z}_i$ is paired with positional embeddings $P_{\mathbf{V}} \in \mathbb{R}^{HW \times D} $ to form the input key's spatial part, and enter the following cross-attention layer.

In the cross-attention layer, the $i$-th actor's classification attention map $\mathcal{A}_i \in \mathbb{R}^{N_c \times HW}$ is constructed as follows:
\DeclareRobustCommand{\svdots}{
  \vbox{%
    \baselineskip=0.33333\normalbaselineskip
    \lineskiplimit=0pt
    \hbox{.}\hbox{.}\hbox{.}%
    \kern-0.2\baselineskip
  }%
}
\begin{equation}
  \mathcal{A}_i  \propto \textnormal{softmax} \left(
      \left( \left[ \begin{array}{ccc@{\hspace{3pt}}|@{\hspace{3pt}}ccc}
             & & & \hhh & P_i^{T} & \hhh \\
            & \boldsymbol{q} & & & \svdots &  \\
            & & & \hhh & P_i^{T} & \hhh
      \end{array}  \right] W_{Q} \right)
      \left( \left[ \begin{array}{c|c} 
         \mathbf{z}_i & P_{\mathbf{V}}
      \end{array}\right] W_{K} 
      \right)^{T}
      \right),
\end{equation}where $W_Q$ and $W_K$ are query and key projection matrices of the cross attention, respectively.
Since $W_Q$ and $W_K$ directly transform the features specific to both actor and class, the attention weight's contributions to each class logit become significantly more diverse than prior methods~\cite{zhao2022tuber, chen2023efficient} where such contributions of attention weights only differ by a scale within different classes.\footnote{See Sec.~A. of the supplementary material for details.}
To be specific,
\begin{equation}
  (\mathcal{A}_i)_{(c, m)}  \propto \textnormal{softmax} \left(
      \left( \nu_{(c, i)}^T W_{Q} \right)
      \left( \eta_{(i, m)}^T W_{K} 
      \right)^{T}
      \right),
    \label{eq:attention_contribution}
\end{equation}where $m \in [1, HW]$ indexes a spatial region, $\nu_{(c,i)}$ is a vector that is conditioned to class and actor, and $\eta_{(i,m)}$ is a vector that is conditioned to actor and region.
As the class index $c$ and the actor index $i$ changes in ~\cref{eq:attention_contribution}, the following variation of the attention map can be trained with the transformer weights dynamically, and thus, we achieve the goal of obtaining both class-specificity and actor-specificity.

The output of CDL $\tilde{\boldsymbol{q}} \in \mathbb{R}^{N_c \times D}$ is subsequently passed to the next layer to serve as class queries again.
To derive the probability for each class, $\tilde{\boldsymbol{q}}$ from the last layer is mean pooled across the channel dimension and then processed through a sigmoid layer.
Thus, the final output $\tilde{\boldsymbol{q}}\in [0,1]^{N_c}$ becomes the classification score for the $i$-th actor.

\subsection{Training objective}

The outputs of the decoder, $\tilde{\boldsymbol{q}}$, $A_i$, and $\mathbf{f}_i$ all contribute to the calculation of the loss.
The confidence score $\hat{\boldsymbol{p}} \in [0,1]^{N_a \times 1}$ for each actor is derived from $\mathbf{f}_i$ to determine the validity of the $i$-th actor's classification and box regression results.
The final box and class outputs $\hat{Y} = \bigl\{ \bigl(\hat{\mathbf{B}}_i, \hat{\mathbf{C}}_i \bigr) | \hat{\mathbf{B}}_i \in \mathbb{R}^{T \times 4}, \hat{\mathbf{C}}_i \in [0, 1]^{T \times N_c}, \ i \in [1, N_a] \bigr\}$ are first processed with Hungarian algorithm~\cite{kuhn1955hungarian} to ensure the model to output a single optimal detection result per actor.
We describe the details of the matching procedure in Sec.~F of the supplementary materials.
The prediction results that are matched with ground truth boxes and classes receive loss signals to output correct answers, while the remaining predictions are trained to output zero probabilities.
Let Y be a padded ground truth labels of size $N_a$, i.e., $Y = \bigl\{\bigl(\mathbf{B}_i, \mathbf{C}_i\bigr) | i \in [1, N_a]\bigr\} $ where $ \mathbf{B}_i = \mathbf{0}_{T \times 4}$ and $ \mathbf{C}_i=\mathbf{0}_{T \times N_c}$ for $i \in [N_X+1, N_a]$. 
Additionally assume that the Hungarian matcher assigns the $i$-th ground truth label of $Y$ to the index $\omega(i)$ of $\hat{Y}$, then each prediction $\hat{Y}_{\omega(i)}$ is passed to the following loss function:
\begin{equation}\label{eqn:eq7}
\begin{split}
    \mathcal{L}(Y_i, \hat{Y}_{\hat{\omega}(i)}) & = 
    \lambda_{\text{class}}\mathcal{L}_{\text{class}}(\mathbf{C}_i, \hat{\mathbf{C}}_{\hat{\omega}(i)}) \\ 
    & + \mathbbm{1}_{i\leq N_X}\lambda_{\text{box}}\mathcal{L}_{\text{box}}(\mathbf{B}_i, \hat{\mathbf{B}}_{\hat{\omega}(i)}) + \mathbbm{1}_{i\leq N_X}\lambda_{\text{giou}}\mathcal{L}_{\text{giou}}(\mathbf{B}_i, \hat{\mathbf{B}}_{\hat{\omega}(i)}) \\
    & + \mathbbm{1}_{i\leq N_X}
    \lambda_{\text{conf}}\mathcal{L}_{\text{conf}}(\mathbf{1}, \hat{\boldsymbol{p}}_{\hat{\omega}(i)}) +
    \mathbbm{1}_{i > N_X}
    \lambda_{\text{conf}}\mathcal{L}_{\text{conf}}(\mathbf{0}, \hat{\boldsymbol{p}}_{\omega(i)}),  \\
\end{split}
\end{equation}
where
\begin{equation}\label{eqn:eq8}
\begin{split}
    \mathcal{L}_{\text{class}}(\mathbf{C}_i, \hat{\mathbf{C}}_{\hat{\omega}(i)}) &= {\text{BFLoss}}(\mathbf{C}_{i}, \hat{\mathbf{C}}_{\hat{\omega}(i)}), \ 
    \mathcal{L}_{\text{box}}(\mathbf{B}_{i}, \hat{\mathbf{B}}_{\hat{\omega}(i)}) = \lVert{\mathbf{B}_{i} - \hat{\mathbf{B}}_{\hat{\omega}(i)}\rVert_{1}}, \\
    \mathcal{L}_{\text{giou}}(\mathbf{B}_{i}, \hat{\mathbf{B}}_{\hat{\omega}(i)}) &= -{\text{GIoU}}(\mathbf{B}_{i}, \hat{\mathbf{B}}_{\hat{\omega}(i)}\bigr), \ 
    \mathcal{L}_{\text{conf}}(\cdot, \hat{\boldsymbol{p}}_{\hat{\omega}(i)}) = {\text{BCELoss}}(\cdot, \hat{\boldsymbol{p}}_{\omega(i)}).
\end{split}
\end{equation}
Notice that BFLoss, BCELoss, and GIoU are a binary focal loss~\cite{lin2017focal}, binary cross-entropy loss, and generalized IoU~\cite{rezatofighi2019generalized}, respectively, and the corresponding lambdas are hyperparameters to balance the loss term.
Since $\mathcal{L}(Y_i, \hat{Y}_{\omega(i)}) \in \mathbb{R}^T$, the final loss $\mathcal{L}$ is represented as:
\begin{equation}\label{eqn:eq9}
    \mathcal{L} = \frac{1}{|T||N_a|}\sum_{t=1}^{T}\sum_{i = 1}^{N_a}{
    \mathcal{L}(Y_i, \hat{Y}_{\hat{\omega}(i)})_{t}}.
\end{equation}
\section{Experiments}
\label{sec:experiments}
\subsection{Experimental setup}
\noindent \textbf{Datasets.}
The model's performance is evaluated on three conventional public benchmarks: 
AVA, JHMDB51-21, and UCF101-24.
AVA~\cite{gu2018ava} is a large scale dataset of 430, 15-minute film/TV show clips.
To be specific, it consists of 211K frames for training and 57K frames for validating.
Due to the large scale of the AVA dataset, it is sparsely annotated at a rate of 1 frame per second (FPS).
Following a standard evaluation protocol, we evaluate the model only at the frame level.
Furthermore, we report the performance based on the refined annotation AVA v2.2.
{JHMDB51-21}~\cite{jhuang2013towards} provides 928 short video clips from YouTube and movie clips.
All videos of JHMDB51-21 are fully annotated with one of 21 actions, and actor's bounding box.
Additionally, videos are trimmed to only contain frames where actions occur, removing the need for temporal localization of the task.
{UCF101-24}~\cite{soomro2012ucf101} provides 3,207 untrimmed YouTube videos, which means it contains frames where no action is taking place.
Therefore, the dataset requires model to be able to distinguish between frames with and without actions.
Following the customary convention, we utilize the corrected annotations~\cite{singh2017online}.

\noindent \textbf{Evaluation criteria.}
The model is evaluated with mean Average Precision (mAP) under IoU threshold of 0.5.
Following the convention, mAP is applied on either a frame-level (f-mAP) or a video-level (v-mAP).

\subsection{Ablation studies}
We carry out ablation studies to justify our choice of design.
A CSN-152~\cite{tran2019video} backbone pretrained on Kinetics-400~\cite{kay2017kinetics} and Instagram65M~\cite{ghadiyaram2019large} is used for the experiments on AVA, and a ViT-B~\cite{dosovitskiy2021an} backbone pretrained on Kinetics-400 is used for the experiments on UCF101-24.

\noindent \textbf{Effectiveness of model components.}
We demonstrate the efficacy of each proposed component of our model.
The baseline configuration of the vanilla model is set to include an ordinary transformer encoder~\cite{vaswani2017attention} and a DETR decoder~\cite{carion2020end}, which are commonly observed baseline structures in recent transformer-based VAD models~\cite{chen2023efficient, zhao2022tuber}.
The impact of each module is assessed through a comparative analysis of the rows in Table~\ref{abl:modules} where the respective modules have been ablated.
As the CDL module is dependent to LDL, we add CDL on top of LDL to measure its effectiveness.
The most prominent improvement comes from the use of 3D Deformable Transformer Encoder and CDL.
Our encoder module takes multi-scale feature maps and enables the model to capture fine details.
Fig.~\ref{ms_ss} illustrates the detection results of the third and fifth row of Table~\ref{abl:modules}, demonstrating the effectiveness of utilizing multi-scale feature maps. 
Still, without the aid of CDL, the advantage of utilizing multi-scale feature maps is not fully maximized: the fourth row shows significantly lower performance than the current model.
In fact, the current state of the art STMixer~\cite{wu2023stmixer} also utilizes multi-scale feature maps, but it underperforms compared to our model (\Cref{main_tab}).

\begin{figure}[t!]
    \begin{minipage}[t]{0.5\textwidth}
        \centering
        \captionof{table}{Ablation experiments on each module of the model.}
        \scalebox{.85}{
        \begin{tabular}{llc}
            \toprule
            {Encoder} & {Decoder} & mAP \\ 
            \hline
            \\[-1em]
            Transformer & DETR & 28.6 \\ 
            Transformer & LDL & 29.1 \\ 
            Transformer & LDL + CDL & 31.4 \\
            3D Deformable Transformer \ & LDL & 31.3 \\
            3D Deformable Transformer \ & LDL + CDL & 33.5 \\
            \bottomrule
        \end{tabular}\label{abl:modules}
        }
        \setcounter{table}{2}
        \captionof{table}{Effect of attaching the actor positional queries to class queries.}
        \scalebox{.90}{
        \begin{tabular}{lcc}
            \toprule
            {Method}  & AVA & UCF \\ 
            \hline
            \\[-1em]
            w/ actor positional queries \qquad \qquad & 33.5 & 85.9 \\
            w/o actor positional queries \qquad \qquad & 31.7 & 82.9 \\
            \bottomrule
        \end{tabular}\label{abl:positional_queries}
        }
    
    \end{minipage}
    \hspace{0.6em}
    \begin{minipage}[t]{0.45\textwidth}
    \centering
    \setcounter{table}{1}
    \captionof{table}{Ablation experiments on ways to aggregate multi-scale feature maps.}
    \scalebox{.9}{
        \resizebox{\textwidth}{!}{
\begin{tabular}{lcc}
    \toprule
    Method \qquad \qquad \qquad  \qquad \qquad  & AVA  & UCF \\ 
    \hline
    \\[-1em]
    actor-specific & 33.5  & 85.9 \\
    weighted sum & 32.9  & 82.0 \\
    mean pooling & 32.0 & 82.8 \\
    \bottomrule
\end{tabular}
}\label{msss}\label{abl:msss}
        }
    \setcounter{table}{3}
    \captionof{table}{Ablation experiments on ways to combine actor and context features.}
    \scalebox{.87}{
        \begin{tabular}{lcc}
            \toprule
            {Method}  & AVA & UCF\\ 
            \hline
            \\[-1em]
            summation & 33.5 & 85.9 \\
            concat + 1d conv~\cite{sun2018actor} \qquad \qquad & 31.8 & 84.7 \\
            cross-attention~\cite{zhao2022tuber} & 31.3 & 81.3 \\
            self-attention~\cite{chen2023efficient} & 30.8 & 81.0 \\
            \bottomrule
        \end{tabular}\label{abl:comb_actor_context}
    }
    \label{msss2}
    \end{minipage}

\end{figure}

\begin{figure}[t!]
    \begin{minipage}[t]{0.453\textwidth}
        \centering
        \includegraphics[width=\textwidth]{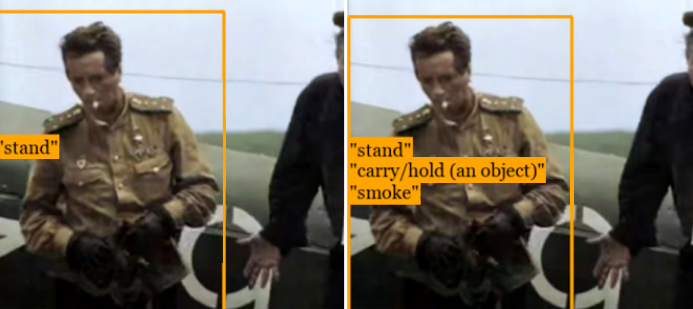}
        \captionof{figure}{{Detection results from the model that uses a single-scale feature map (left) and multi-scale feature maps (right).} Detection threshold is set to 0.5.}
        \label{ms_ss}
    \end{minipage}    
    \hfill
    \begin{minipage}[t]{0.525\textwidth}
        \centering
        \includegraphics[width=\textwidth]{
        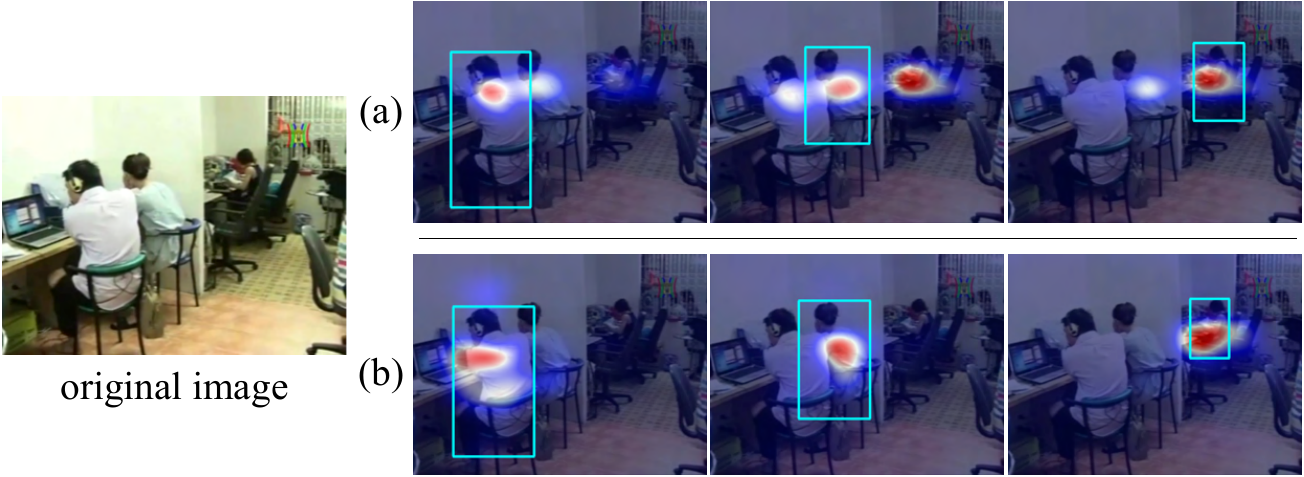}
        \captionof{figure}{Attention visualizations of the class, \textsl{`sit'}.
        (a) shows the model without the attachment of the actor positional queries to class queries, and (b) depicts the model with this attachment.}
        \label{wopq}
    \end{minipage}

\end{figure}

\noindent \textbf{Different ways to aggregate multi-scale feature maps.}
In LDL, we aggregate multi-scale feature maps into single-scale feature maps, while conditioning them to each actor to provide actor-specific context to CDL.
Actor-conditioned aggregation helps identifying context features that are relevant to the actor of interest and mitigates the issue of mistakenly gathering context that is relevant to same actions of different actors.
\Cref{msss} shows its effectiveness: providing actor-specific context demonstrates superior results on both datasets.

\noindent \textbf{Effectiveness of using the actor positional queries for classification.}
As explained in Sec.~\ref{sec:Classfier}, we attach the actor positional queries to class queries in order to provide class queries with information regarding the actor the model aims to classify.
Similarly to actor-conditioned aggregation from the previous section, the attachment of the actor positional queries improves actor-specificity of the class queries and prevents the activation of the context that belongs to wrong actors.
Fig.~\ref{wopq}(a) shows the case where the actor positional queries are not attached to class queries: class queries activate regions where a similar actor performs the same action.
In contrast, Fig.~\ref{wopq}(b) illustrates the impact of the actor positional queries, mitigating the issue of actor-agnostic activation.
The results from \Cref{abl:positional_queries} also support our claim: detaching the actor positional queries compromises the performance by a significant margin.

\noindent \textbf{Different ways to combine actor and context features.}
Merging the actor and context feature is an essential step in VAD to model the interactions between the actor and the background.
We explore various methods to merge the actor and context features to justify our chosen approach. Each method we chose in Table~\ref{abl:comb_actor_context} corresponds to a way used in prior approaches \cite{sun2018actor, zhao2022tuber, chen2023efficient}. 
Interestingly, we discover that simple summation yields superior performance overall.

\subsection{Comparison to the state-of-the-art methods}

\setlength{\floatsep}{0pt plus 0pt minus 10pt}
\setlength{\textfloatsep}{0pt plus 0pt minus 5pt}
\begin{table*}[t!]
\centering
\caption{Performance comparison on three benchmarks~\cite{gu2018ava, jhuang2013towards, soomro2012ucf101}. The column labeled ``\emph{D}'' signifies whether a model utilizes an off-the-shelf detector or not. 
}
\subfloat[Performance comparison on AVA v2.2.]{\label{main_tab:tab1a}\scalebox{.734}{
\scalebox{.82}{
\begin{tabular}{lccccccc}
\toprule
Model                                        & \:\emph{D}\:                & Input        & Backbone      & Pre-train   & mAP \\ 
\hline
\\[-1em]
SlowFast~\cite{feichtenhofer2019slowfast}      &\cmark                   &  $32 \times 2$ & SF-R101-NL    & K600          & {29.0}  \\ 
ACAR~\cite{pan2021actor}                       &\cmark                   &  $32 \times 2$ & SF-R101-NL    & K600          & {31.4}  \\ 
AIA~\cite{tang2020asynchronous}                &\cmark                   &  $32 \times 2$ & SF-R101       & K700          & {32.3}  \\ 
VideoMAE~\cite{tong2022videomae}               &\cmark                   &  $16 \times 4$ & ViT-B         & K400          & {31.8}  \\ 
MeMViT~\cite{wu2022memvit}                     &\cmark                   &  $32 \times 3$ & MViTv2-B      & K600          & {32.8}  \\ 
WOO~\cite{chen2021watch}                       & \xmark                  &  $32 \times 2$ & SF-R101-NL    & K600          & {28.3}  \\ 
TubeR~\cite{zhao2022tuber}                     & \xmark                  &  $32 \times 2$ & CSN-152       & IG65M, K400   & {31.1}  \\ 
STMixer~\cite{wu2023stmixer}                   & \xmark                  &  $32 \times 2$ & CSN-152       & IG65M, K400   & \underline{32.8}  \\ 
STMixer~\cite{wu2023stmixer}                   & \xmark                  &  $16 \times 4$ & ViT-B         & K400          & \underline{32.6}  \\ 
STMixer~\cite{wu2023stmixer}                   & \xmark                  &  $16 \times 4$ & ViT-B~\cite{wang2023videomae}         & K710, K400    & {36.1}  \\
EVAD~\cite{chen2023efficient}                  & \xmark                  &  $16 \times 4$ & ViT-B         & K400          & {32.3}  \\
EVAD~\cite{chen2023efficient}                  & \xmark                  &  $16 \times 4$ & ViT-B~\cite{wang2023videomae}         & K710, K400    & \underline{37.7}  \\
\rowcolor{Gray!20}
Ours                                           & \xmark                  &  $32 \times 2$ & CSN-152       & IG65M, K400   & {\textbf{33.5}}  \\ 
\rowcolor{Gray!20}
Ours                                           & \xmark                  &  $16 \times 4$ & ViT-B         & K400          & {\textbf{32.9}}  \\
\rowcolor{Gray!20}
Ours                                           & \xmark                  &  $16 \times 4$ & ViT-B~\cite{wang2023videomae}         & K710, K400    & {\textbf{38.4}} \\

\bottomrule
\end{tabular}
}
}}
\hfill
\subfloat[Performance comparison on JHMDB and UCF.]{\label{main_tab:tab1b}\scalebox{.72}{

\scalebox{.83}{
\begin{tabular}{lccccccc}
\toprule
\multirow{2}{*}{Model}
& \multirow{2}{*}{$\:\emph{D}\:$}
& \multicolumn{2}{c}{Input}
& \multirow{2}{*}{Backbone}
& \multicolumn{2}{c}{f-mAP / v-mAP} \\
\cmidrule{3-4} \cmidrule{6-7}
& & JHMDB & UCF
& & JHMDB & UCF \\[0.5mm]
\hline
\\[-1em]
MOC~\cite{li2020actions}            &\cmark  & \multicolumn{2}{c}{$7 \times 1$}   & DLA34           & 70.8 / 77.2   & 78.0 / 53.8  \\ 
AVA~\cite{gu2018ava}                &\xmark  & \multicolumn{2}{c}{$20 \times 1$}  & I3D-VGG         & 73.3 / 78.6   & 76.3 / \underline{59.9}  \\ 
ACRN~\cite{sun2018actor}            &\xmark  &  $20 \times 1$ &      -            & SF-R101         & 77.9 / 80.1   & {- \:\:\:/ \:\:\:-}  \\ 
CA-RCNN~\cite{wu2020context}        &\cmark  &  $32 \times 2$ &      -            & R50-NL          & 79.2 / { \:\:- \:\:} & {- \:\:\:/ \:\:\:-}  \\ 
YOWO~\cite{kopuklu2019you}          &\xmark  & \multicolumn{2}{c}{$16 \times 1$}  & 3DResNext-101   & 74.4 / \underline{85.7}   & 80.4 / 48.8 \\
WOO~\cite{chen2021watch}            &\xmark  &  $32 \times 2$ &      -            & SF-R101-NL      & 80.5 / { \:\:- \:\:} & { \:\:- \:\:} / { \:\:- \:\:}  \\ 
AIA~\cite{tang2020asynchronous}     &\cmark  &  -             & $32 \times 1$     & R50-C2D         & {- \:\:\:/ \:\:\:-}           & 78.8 / { \:\:- \:\:} \\ 
ACAR~\cite{pan2021actor}            &\cmark  &  -             & $32 \times 1$     & SF-R50          & {- \:\:\:/ \:\:\:-}          & 84.3 / { \:\:- \:\:} \\ 
TubeR~\cite{zhao2022tuber}          &\xmark  & \multicolumn{2}{c}{$32 \times 2$}  & CSN-152         & { \:\:- \:\:} / 82.3 & 83.2 / 58.4 \\ 
STMixer~\cite{wu2023stmixer}        &\xmark  & \multicolumn{2}{c}{$32 \times 2$}  & SF-R101-NL      & 86.7 / { \:\:- \:\:} & 83.7 / { \:\:- \:\:} \\ 
EVAD~\cite{chen2023efficient}       &\xmark  & \multicolumn{2}{c}{$16 \times 4$}  & ViT-B           & {\textbf{90.2}} / 77.8 & \underline{85.1} / 58.8 \\ 
\rowcolor{Gray!20}
Ours                               &\xmark  &  $40 \times 1$ & $32 \times 1$      & ViT-B           & \underline{86.9} / {\textbf{88.5}}  & {\textbf{85.9}} / {\textbf{61.7}} \\ 
\bottomrule
\end{tabular}
}

}}
\label{main_tab}
\end{table*}

Table~\ref{main_tab:tab1a} summarizes the experiment results on AVA v2.2. 
We compare our model with recent video backbones widely used in video action detection, ViT-B~\cite{dosovitskiy2021an, wang2023videomae} and CSN-152~\cite{tran2019video}.
Among the models that utilize ViT-B as their backbones, our model surpasses other models~\cite{wu2023stmixer, chen2023efficient} that have identical settings to ours.
Out of the models employing CSN-152 as well our model exhibits a superior performance, highlighting its competitive edge.

\setlength{\floatsep}{0pt plus 0pt minus 10pt}
\setlength{\textfloatsep}{12pt plus 0pt minus 5pt}
\begin{figure}[t!]
    \begin{minipage}[t]{0.59\textwidth}
    \centering
    \captionof{table}{Efficiency comparison on JHMDB and UCF. Each frame's resolution is set to 256 $\times$ 360.
    }
        \scalebox{0.63}{
        \begin{tabular}{>{\raggedright\arraybackslash}m{1.2cm}>{\centering\arraybackslash}m{2.3cm}>{\centering\arraybackslash}m{1.2cm}>{\centering\arraybackslash}m{1.2cm}>{\centering\arraybackslash}m{1.2cm}>{\centering\arraybackslash}m{1.2cm}>{\centering\arraybackslash}m{2.4cm}>{\centering\arraybackslash}m{3.8cm}}
        \toprule
        {Method} & {Backbone} & {Input $(T \times \tau)$ }      & {Params}  & {FLOPs}  & {\scriptsize Inf. time (ms)}  & {f-mAP/v-mAP}  \\ 
        \hline
        \\[-1em]
        \multicolumn{7}{l}{\textbf{JHMDB}: inferring a 40-frame tube}\\
        \hline
        \\[-1em]
        TubeR    & CSN-152  & $32 \times 2$  & \textbf{91.8M}   & 6.99T  & 1263.1  & \:\:-\:\: / \underline{82.3} \\
        STMixer  & SF-R101-NL & $32 \times 2$ & 219.2M   & 7.64T  & 2088.2  & 86.7 / \:\:-\:\: \\
        EVAD     & ViT-B    & $16 \times 4$ & 185.4M  & 10.68T & 8363.1 & \textbf{90.2} / 77.8 \\
        \rowcolor{Gray!20}
        Ours     & ViT-B    & $40 \times 1$ & \underline{117.8M}  & \textbf{3.26T}  & \textbf{432.0}  & {\underline{86.9}} / \textbf{88.5} \\
        \hline
        \\[-1em]
        \multicolumn{7}{l}{\textbf{UCF}: inferring a 32-frame tube}\\
        \hline
        \\[-1em]
        TubeR    & CSN-152  & $32 \times 2$ & \textbf{91.8M}   & 5.60T  & 1161.9 & 83.2 / 58.4 \\
        STMixer  & SF-R101-NL & $32 \times 2$ & 219.2M  & 6.12T  & 1671.5 & 83.7 / \:\:-\:\: \\
        EVAD     & ViT-B    & $16 \times 4$ & 185.4M  & 8.54T  & 6797.8 & \underline{85.1} / \underline{58.8} \\
        \rowcolor{Gray!20}
        Ours     & ViT-B    & $32 \times 1$ & \underline{117.8M}  & \textbf{3.73T}  & \textbf{370.0}  & \textbf{85.9} / \textbf{61.7} \\        
        
        \bottomrule
        \end{tabular}
        }
        \label{flops}
    \end{minipage}
    \hfill
    \begin{minipage}[t]{0.38\textwidth}
        \centering
        \captionof{table}{Performance comparison on the latest VAD models.}
        \scalebox{0.77}{
        \begin{tabular}{lccc}
        \toprule
        \multirow{2}{*}{Method} & \multirow{2}{*}{Backbone} & \multirow{2}{*}{f-mAP} & when GT box \\ 
         &  &  & is provided \\ [0.2em]
        \hline
        \\[-1em]
        TubeR & CSN-152 & 31.1 & $33.1_{(+2.0)}$ \\
        STMixer & ViT-B & 32.6 & $34.7_{(+2.1)}$ \\
        STMixer & ViT-B~\cite{wang2023videomae} & 36.1 & $38.7_{(+2.6)}$ \\
        EVAD & ViT-B & 32.3 & $34.5_{(+2.2)}$ \\
        EVAD & ViT-B~\cite{wang2023videomae} & 37.7 & $40.0_{(+2.3)}$ \\
        \rowcolor{Gray!20}
        Ours & CSN-152 & \textbf{33.5} & $\mathbf{37.2_{(+3.7)}}$ \\
        \rowcolor{Gray!20}
        Ours & ViT-B & \textbf{32.9} & $\mathbf{36.9_{(+4.0)}}$ \\
        \rowcolor{Gray!20}
        Ours & ViT-B~\cite{wang2023videomae} & \textbf{38.4} & $\mathbf{42.1_{(+3.7)}}$ \\

        \bottomrule
        \end{tabular}
        }
        \label{classification metric 2}
    \end{minipage}
\end{figure}

Table~\ref{main_tab:tab1b} shows the results on JHMDB51-21 and UCF101-24.
Our model demonstrates comparable or superior performance to current state-of-the-art models on both datasets.
We conjecture the lower performance compared to the prior method~\cite{chen2023efficient} is due to the low class diversity of JHMDB51-21, since the strength of our model comes from its ability to create diverse classification feature conditioned to each action class.

Moreover, the tube-creating property of our model largely improves the model's efficiency.
Table~\ref{flops} shows the amount of computation utilized by current models to generate a single tube.
Note that the input size $T \times \tau$ signifies the number of frames $T$ and sampling rate $\tau$.
The current best-performing models~\cite{chen2023efficient, wu2023stmixer} produce predictions only for a single frame within a single feedforward, while ours produce a whole tube at once.
Thus, although the input size differs, taking the input with larger $T$ with sampling rate 1 is to infer a tube with a single feedforward, thereby compensating for the multi-feedforwarding property of other models that process smaller input sizes, thus ensuring fairness in comparison.
Specificially, for JHMDB51-21, with its longest clip comprising 40 frames, we set $T=40$, while for UCF101-24, where the longest clip spans 900-frame long, thus we adopt $T=32$, following the standard practice.
Accordingly, we measure the amount of computation and speed needed for generating a 40-frame tube for JHMDB51-21 and a 32-frame tube for UCF101-24.
Notably, our model achieves competitive results with fewer parameters and the least FLOPs.
Due to the page limit, we provide a detailed explanation of the model's efficiency in Sec.G of the supplementary material.

\subsection{Attention visualization and analysis}

\begin{figure*}[t!]
     \centering
     \includegraphics[width=0.98\textwidth]{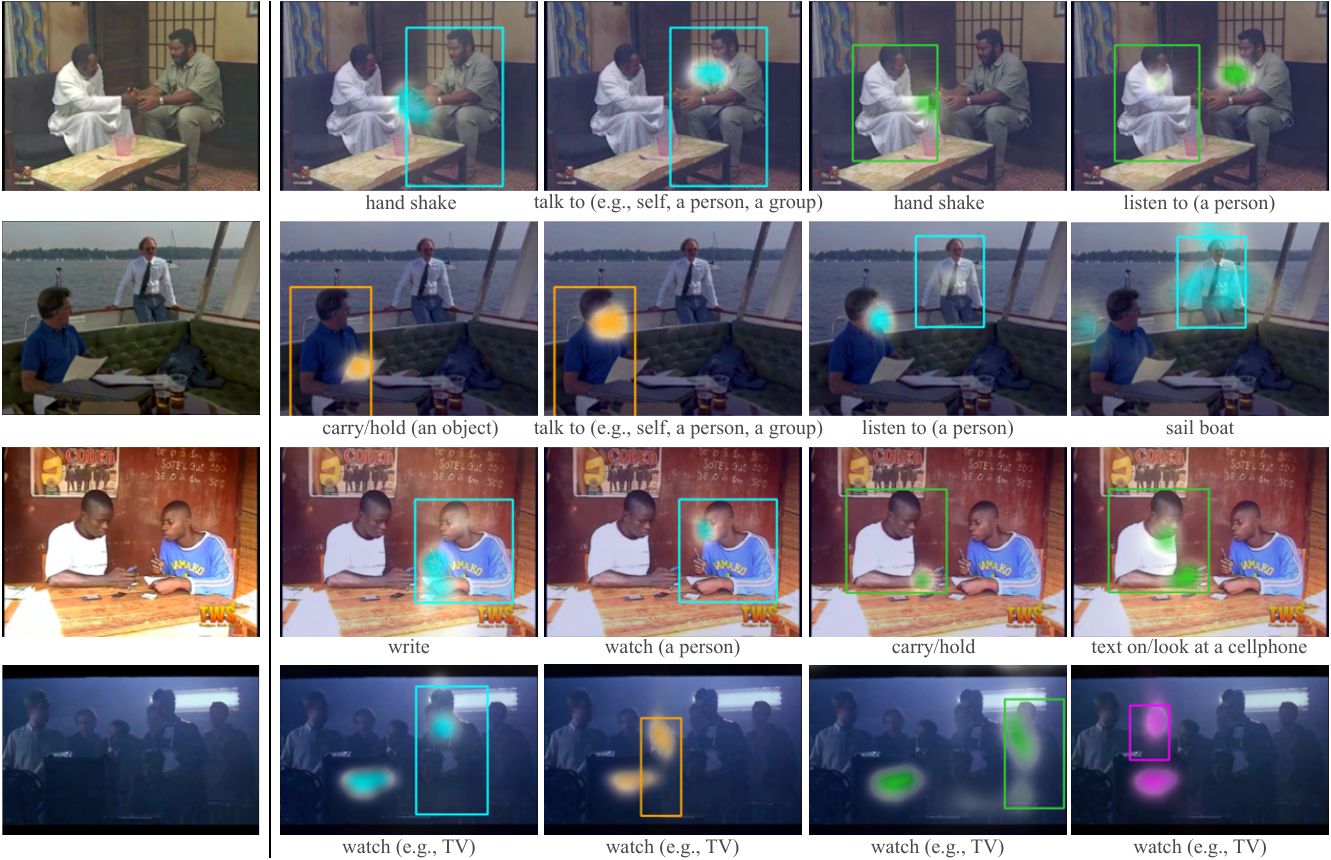}
     \caption{Classification attention map visualization results on AVA v2.2. Each map represents the region where the model has observed to classify the action of the actor, marked with the same color as the bounding box. 
     }
     \label{visualization}
\end{figure*}

One of key advantages of our model is its ability to provide interpretable attention maps for individual class labels.
Fig.~\ref{visualization} shows the classification attention maps the model provides across various scenarios.
The model dynamically focuses on crucial regions; for example, when classifying actions like \textsl{`hand shake'} or \textsl{`write,'} it tends to examine the areas near the actor's hands and arms.
If necessary, the model also directs its attention towards regions outside the actor's body such as \textsl{`listen to (a person)'} in the first and second row, or \textsl{`watch (e.g., TV)'} in the fourth row.
Moreover, it differentiates class-relevant context of actors well in multi-actor scenarios, even when multiple actors are performing identical actions.

\noindent \textbf{Is performance improvement really from classification?}
Since average precision (AP) simultaneously measures a model's localization and classification ability, we conduct experiments to measure the model's classification ability explicitly (\Cref{classification metric 2}).
The model outputs are Hungarian-matched with the ground-truth (GT) instances using the same matching cost, and their predicted box coordinates are replaced with that of GT annotations.
The experiment results show that the performance gap between ours and other methods has increased, implying that our model's performance improvement is mainly due to its enhanced classification ability.

\section{Conclusion and limitations}
\label{sec:conclusion}
We have introduced a model designed for effective classification, constructing features dedicated to each class and each actor.
Our approach adeptly manages classification features by considering context not bounded by the actor's location, while still capturing information specific to the targeted actor. 
We achieve the state-of-the-art performance on the most challenging benchmark, while being the most efficient network when inferring a tube.

\noindent \textbf{Limitations.}
However, there still exists room for further improvement.
The current decoder architecture does not exchange information across frames due to the restricted size of memory.
As a result, a role of capturing temporal dynamics is solely dependent on the encoder architecture, placing a significant burden on the encoder.
Thus, further studies might aim to achieve greater performance by sparsely collecting class information from the spatial domain, leaving residual memory available for capturing temporal dynamics.

\newpage
\title{\suppPaperTitle}
\titlerunning{\abbvPaperTitle}

\author{\authorBlock}
\authorrunning{\abbvAuthors}
\institute{\authorInstitutes}

\maketitle

\appendix
\renewcommand\thefigure{A\arabic{figure}}
\renewcommand{\thetable}{A\arabic{table}}
\renewcommand{\theequation}{A\arabic{equation}}
\setcounter{figure}{0}
\setcounter{table}{0}
\setcounter{equation}{0}


We present omitted experiments and details in this supplementary material as below:
\begin{itemize}
    \item Sec.~\ref{appendix: attn_map_contribution}: Analyzing Previous Models' Attention Weight Contributions to Classification Logits
    \item Sec.~\ref{appendix:single-label}: Implementation for Single-label Datasets
    \item Sec.~\ref{appendix:3DDefEnc}: Detailed Logic of the 3D Deformable Encoder
    \item Sec.~\ref{appendix:Details_LDL}: Detailed Logic of Localizing Decoder Layer (LDL)
    \item Sec.~\ref{appendix:Details_CDL}: Detailed Logic of Classifying Decoder Layer (CDL)
    \item Sec.~\ref{appendix:Hungarian_matching}: Details of the Hungarian Matching Process
    \item Sec.~\ref{appendix:model_comparison}: Comparison with Latest VAD Models    
    \item Sec.~\ref{appendix:experimental_setup}: Experimental Setup    
    \item Sec.~\ref{appendix:additional_ablations}: Additional Ablation Experiments
    \item Sec.~\ref{appendix:more_qualitative}: Additional Qualitative Results    
    \item Sec.~\ref{appendix:classwise_map}: Class-wise mAP Comparison
    
\end{itemize}


\section{Analyzing Previous Models' Attention Weight Contributions to Classification Logits}\label{appendix: attn_map_contribution}
To further investigate the difference between prior transformer-based methods~\cite{zhao2022tuber, chen2023efficient} and our model, we analyze how attention weights of the previous models affect the final classification logits.
In summary, we find that prior methods allow their each class logit to have comparably subtle differences within different classes with respect to the acquired attention weights, and thus, enforce their transformer outputs to include more commonly shared semantics across distinct classes.

\begin{figure}[t!]
     \begin{subfigure}[b]{0.45\textwidth}
         \centering
         \includegraphics[width=\textwidth]{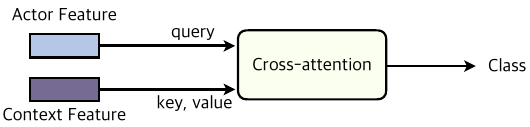}
         \caption{Classification module of TubeR~\cite{zhao2022tuber}}
         \label{fig:prev_1}
     \end{subfigure}
     \hspace{3mm}
     \begin{subfigure}[b]{0.49\textwidth}
         \centering
         \includegraphics[width=\textwidth]{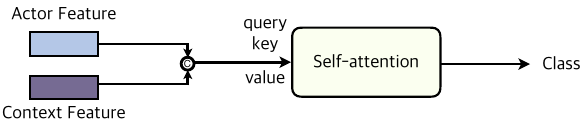}
         \caption{Classification module of EVAD~\cite{chen2023efficient}}
         \label{fig:prev_2}
     \end{subfigure}
     \caption{Two previous approaches for action classification using transformer.
     }
     \vspace{-5mm}
     \label{appendix:prev_methods}
\end{figure}

In Fig.~\ref{appendix:prev_methods}, we present how previous methods construct their classification features.
The notations defined in Sec.~4 of the main paper are adaptively reused for a clearer comparison to our method:
we denote the actor feature as $\mathbf{f} \in \mathbb{R}^{N_a \times D}$ and the context feature as $\mathbf{V} \in \mathbb{R}^{THW \times D}$.
TubeR~\cite{zhao2022tuber} constructs its classification feature with a cross-attention layer whose input query is $\mathbf{f}$ and input key/value are $\mathbf{V}$.
Then, the $i$-th actor's classification attention map $\mathcal{A}$ is constructed as follows:
\DeclareRobustCommand{\svdots}{
  \vbox{%
    \baselineskip=0.33333\normalbaselineskip
    \lineskiplimit=0pt
    \hbox{.}\hbox{.}\hbox{.}%
    \kern-0.2\baselineskip
  }%
}
\begin{equation}\label{eqn:tuber_attnmap}
  \mathcal{A}_i  \propto
  \textnormal{softmax} \left(
  \left(
      \mathbf{f}_i^{T} W_{Q} \right)
      \left( \mathbf{V} W_{K} 
      \right)^{T}
      \right),
\end{equation}where $W_Q$ and $W_K$ are $D$ by $D$ projection matrices of the transformer.
Since the same attention map is shared across to infer different classes, the final linear layer, which we will denote its parameter as $W^{\text{cls}} \in \mathbb{R}^{N_c \times D}$, needs to be involved to further analyze the impact of attention weights to the classification logits.
For brevity, we disregard the bias term in this discussion.
Let $c$-th class logit of the $i$-th actor be $\ell_{(i,c)}$, then it is derived from Eq.~\ref{eqn:tuber_attnmap} as:
\DeclareRobustCommand{\svdots}{
  \vbox{%
    \baselineskip=0.33333\normalbaselineskip
    \lineskiplimit=0pt
    \hbox{.}\hbox{.}\hbox{.}%
    \kern-0.2\baselineskip
  }%
}
\begin{equation}\label{appendix:tuber_eqn}
\begin{split}
  \mathcal{\ell}_{(i, c)}  &\propto
  W^{\text{cls}}_{c}
  \left( \sum_{m}
  \left(
    \mathcal{A}_{(i, m)}
    \odot \mathbf{V}_m W_{V}
  \right)
  \right) \\
  & = 
  \sum_{d} \left( W^{\text{cls}}_{(c,d)}
  \sum_{m} \nu_{(i,m,d)}
  \right),
\end{split}
\end{equation}where $m \in [1, THW], d \in [1, D], W_V$ is a projection matrix, and $\nu_{(i,m, d)}$ is a scalar that is conditioned to actor, region and channel dimension.
As the class index $c$ varies in Eq.~\ref{appendix:tuber_eqn}, the only relevant factor that influences the impact of the attention weights on classification logits is $W^{\text{cls}}_{(c,d)}$, which is merely a scalar value that varies across different classes.
Although slightly different, such impact is similarly derived in the case of EVAD~\cite{chen2023efficient}.
Therefore, the weight responsible for distinguishing distinct classes is solely dependent on the elements of $W^{\text{cls}}$.
Given the limited variation that can be generated within classes at the final layer, the transformer weights are trained to alleviate the burden of $W^{\text{cls}}$ by capturing the commonly shared semantics (\ie, actors) across different classes, so that each action class can be expressed with a unique combination of such semantics.

On the other hand, our model involves class-specificity during the generation of attention weights, and thus, the distinction between action classes no longer depends on the final linear layer.
This effectively enhances the expressiveness of the classification feature, eventually aiding in solving VAD, which particularly requires the identification of subtle differences between individual action classes.

\section{Implementation for single-label datasets}\label{appendix:single-label}
The description of Sec.~4 of the main paper assumes that the given dataset has a multi-label property, since it covers more general scenarios that include the cases of single-label datasets.
Still, we slightly change the model's implementation to inject this single-label prior.
To be specific, since the ground-truth label $\{\mathbf{C}_i \in \{0, 1\}^{T \times N_c} | \ i \in [1, N_X]\}$ is a one-hot vector for each timestep, the model's classification output $\{\hat{\mathbf{C}}_i \in [0, 1]^{T \times N_c} | \ i \in [1, N_X]\}$ is under the constraint of $\sum_{c=1}^{N_c}{\hat{\mathbf{C}}_i[t,c]} = 1 \ \forall i \in [1, N_X], \forall t \in [1, T]$.
Thus, the output $\tilde{\boldsymbol{q}} \in \mathbb{R}^{N_c \times 1}$ (after mean pooling) undergoes through a softmax layer instead of the sigmoid layer.

Furthermore, classification outputs that are not matched with GT labels, i.e., $\hat{\mathbf{C}}_{\omega(i)} \forall i > N_X$, are trained to output zero probabilities for all classes.
However, it is not possible if the output is processed through the softmax layer.
Thus, we additionally involve the confidence score $\hat{\boldsymbol{p}}_{i} \in [0,1]$ of each actor candidate to refine the classification logits, so that the unmatched outputs can return zero vector if necessary.

\section{Detailed logic of 3D Deformable Transformer Encoder}\label{appendix:3DDefEnc}

Due to the heavy nature of multi-scale spatio-temporal feature maps, we adopt ideas from Deformable DETR~\cite{zhu2020deformable} for an efficient encoding process.
We extend the idea to incorporate temporal dynamics, since the original approach only allows the information exchange within a single frame.
Note that every $\boldsymbol{v} \in \{ \boldsymbol{v}^l(t,h,w) \rvert (t,h,w) \in {[1, T_l] \times [1, H_l] \times [1, W_l]} \}_{l=1}^{L} \subset \mathbb{R}^D$ is regarded as the query and goes through the identical encoding process, where
$\boldsymbol{v}^l(t,h,w)$ indicates the 1D feature located in the $(t,h,w)$-th position in $\boldsymbol{v}^l$.
Let $q$ index a query element of the encoder where its normalized coordinates and its corresponding query feature are denoted by $\hat{\boldsymbol{p}}_{q} \in [0,1]^3$ and $\boldsymbol{v}_{q} \in \mathbb{R}^D$, respectively.
Given $\mathbf{V} = \{\boldsymbol{v}^l \}_{l=1}^{L}$, the 3D multi-scale deformable encoder module is applied for each $\boldsymbol{v}_{q}$ as
\begin{equation}\label{aeqn:eq1}
\begin{split}
    {\textnormal{3DMSDeformableEncoder}\Bigl({ \boldsymbol{v}_{q}, \hat{\boldsymbol{p}}_{q}, \{ \boldsymbol{v}^{l} \}_{l=1}^{L} }\Bigr) } & \\ 
    =\sum_{m=1}^{M} \boldsymbol{W}_{m} \Bigl[ \sum_{l=1}^{L}{\sum_{k=1}^{K} {A_{mlqk}}} &
    \cdot \boldsymbol{W}_{m}' \boldsymbol{v}^l \Bigl( {\phi}_{l}(\hat{\boldsymbol{p}}_{q}) + \Delta \boldsymbol{p}_{mlqk} \Bigr ) \Bigr ],
\end{split}
\end{equation}
where $m$ indexes the attention head, $l$ indexes the input feature level, and $k$ indexes the sampling point.
Similar to the previous work, $\boldsymbol{W}_m' \in \mathbb{R}^{(D/M) \times D}$ and $\boldsymbol{W}_m \in \mathbb{R}^{(D/M) \times D}$ are learnable weights that weigh across multiple attention heads.
Also, the offset $\Delta\boldsymbol{p}_{mlqk}  \in \mathbb{R}^3$ and the weight $A_{mlqk} \in \mathbb{R}$, which is dedicated to each $k$-th sampling point at the $l$-th level of the feature map for the $q$-th query's $m$-th attention head, are obtained by applying linear projection layers to $\boldsymbol{z}_q$.
Note that $\phi_l(\cdot)$ re-scales the coordinates to the $l$-th level input feature map, and $\sum_{k=1}^K{A_{mlqk}} = 1$.

\section{Detailed logic of Localizing Decoder Layer (LDL)}\label{appendix:Details_LDL}
\begin{figure*}[t!]
     \centering
     \includegraphics[width=0.9\textwidth]{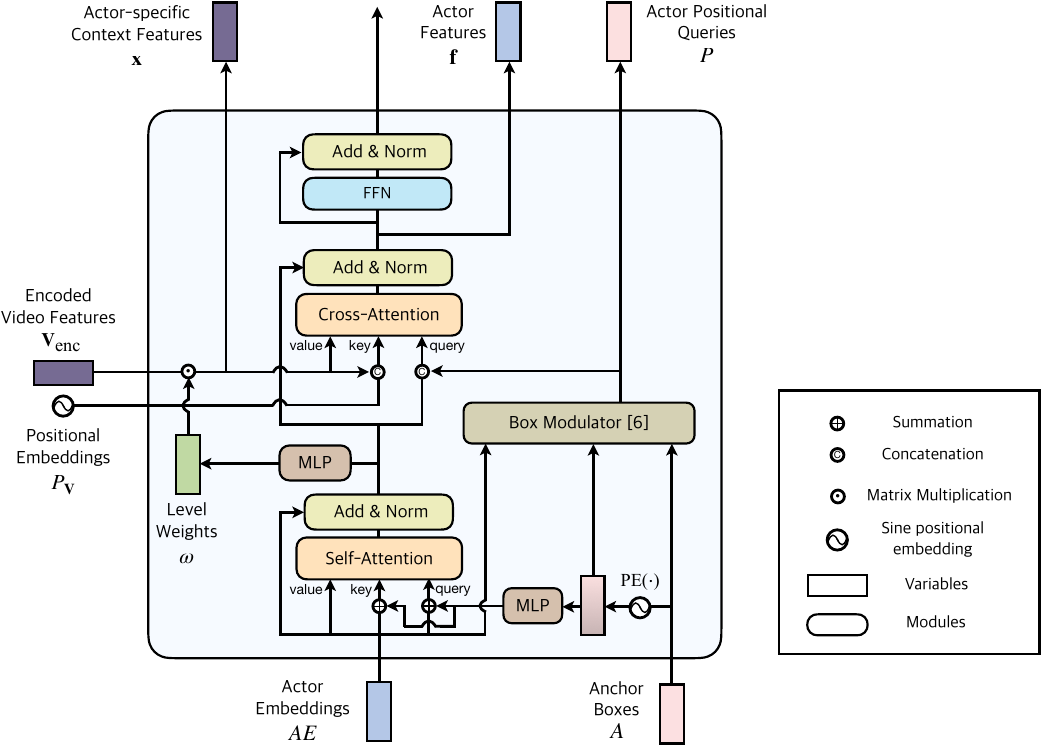}
     \caption{Structure of Localizing Decoder Layer (LDL)}
     \label{LDL_detailed}
     \vspace{-1mm}
\end{figure*}




In this section, we provide details of Localizing Decoder Layer (LDL), of which we omit the details in the main paper.
Fig.~\ref{LDL_detailed} illustrates the transformation process of LDL.
Let the $i$-th actor box and zero-initialized actor embedding be $A_i = (x_i, y_i, w_i, h_i)$ and $AE_i$, respectively, where
$i \in [1, N_a]$ and $N_a$ is the number of actor candidates that are each assigned to capture a potential actor that appears in the video.
$AE_i \in \mathbb{R}^{T \times D}$ is processed in a temporally parallel manner to output prediction per frame.
Hence, for brevity, we omit the time index $t$ and focus on how an individual frame is processed within the decoder layer.



$AE$ initially passes through a self-attention layer to ensure that each $i$-th actor embedding represents different actors.
The key and value are first embedded with positional information provided by the actor box $A$, and then the $AE$ is updated with self-attention.
Before concatenating the actor boxes to the updated $AE$, the boxes undergo modulation~\cite{liu2022dab} to incorporate more precise position and size information of actors.
This utilizes $AE$ (before the self-attention), which is trained to contain the actor's information.
To be specific, the box modulation BM is applied to each $i$-th element of $A$ as follows:
\begin{equation}\label{aeqn:eq4}
    \begin{split}
    P_{i} &= {\textnormal{BM}\biggl(\Bigl[\textnormal{PE}(x_{i}), \textnormal{PE}(y_{i})\Bigl]; \textnormal{AE}_{i}, A_{i} \biggr)} \\
     &= \Bigl[\textnormal{PE}(x_{i})\frac{w_{i, \textnormal{ref}}}{w_{i}}, \textnormal{PE}(y_{i})\frac{h_{i, \textnormal{ref}}}{h_{i}} \Bigr], \\
    \end{split}     
\end{equation}
where  $(w_{i, \textnormal{ref}}, h_{i, \textnormal{ref}})$ is obtained by applying a fully connected layer to $AE{_i}$, and PE$: \mathbb{R} \rightarrow \mathbb{R}^{D/2} $ is a sinusoidal positional encoding.

Note that in the cross-attention layer, the role of the modulated actor box $P$ is to guide the actor embedding $AE$ to gather information relevant to the $i$-th instance from the encoded feature maps $\mathbf{V}_{\textnormal{enc}}$.
Accordingly, we compose actor-specific context features $\mathbf{x} \in \mathbb{R}^{N_a \times T \times H \times W \times D}$, where $\mathbf{x}_i$ is a feature map dedicated to the $i$-th actor.
Specifically, we apply the weighted summation on $\mathbf{V}_{\textnormal{enc}}$ while conditioning the weights to $AE_i$:
\begin{equation}\label{aeqn:eq5}
    {\textbf{x}_{i} = \sum_{l=1}^L \omega_i^l\textbf{v}_{\textnormal{enc}}^l  \in \mathbb{R}^{T \times H \times W \times D}}, \\
\end{equation}    
where $(\omega_i^1, \omega_i^2, \ldots, \omega_i^{L}) = \textnormal{MLP}(AE_i).$
Constructing distinct values that are specific to each actor enhances actor-specificity, helping the model to generate different features for each actor.
Afterwards, 
the cross-attention integrates the features composed so far as follows:
\begin{equation}\label{aeqn:eq6}
    {AE_i \leftarrow f_{\text{CA}}^{\textnormal{loc}} \biggl(\textsl{q = }\Bigl[{AE_i}, {P_i}\Bigr],} \
    {\textsl{k = }\Bigl[\textbf{x}_{i}, P_\textbf{V}\Bigr],} \
    {\textsl{v = }\textbf{x}_{i}\biggl),}
\end{equation}
%
where ${P_i}$ is from Eq.~\ref{aeqn:eq4} and $P_\textbf{V} \in \mathbb{R}^{T \times H \times W \times D}$ is a 3D positional embedding of the video feature maps $\mathbf{V}$.

Before the output $AE$ undergoes a feedforward network, it is passed to the CDL, and we denote this vector $\mathbf{f}$ as an actor feature.
Afterwards, $AE$ contributes to refine the anchor box $A$ as follows:
\begin{equation}\label{aeqn:eq7}
A \leftarrow \sigma(\textnormal{FFN}(AE) + \sigma^{-1}(A)),
\end{equation}
where FFN($\cdot$), $\sigma: \mathbb{R} \rightarrow [0,1]$ and $\sigma^{-1}: [0,1] \rightarrow \mathbb{R}$ are a feedforward network, sigmoid function, and its inverse that normalizes and unnormalizes the input, respectively.

\section{Details of the Classifying Decoder Layer (CDL)}\label{appendix:Details_CDL}

\begin{figure*}[t!]
     \centering
     \includegraphics[width=0.9\textwidth]{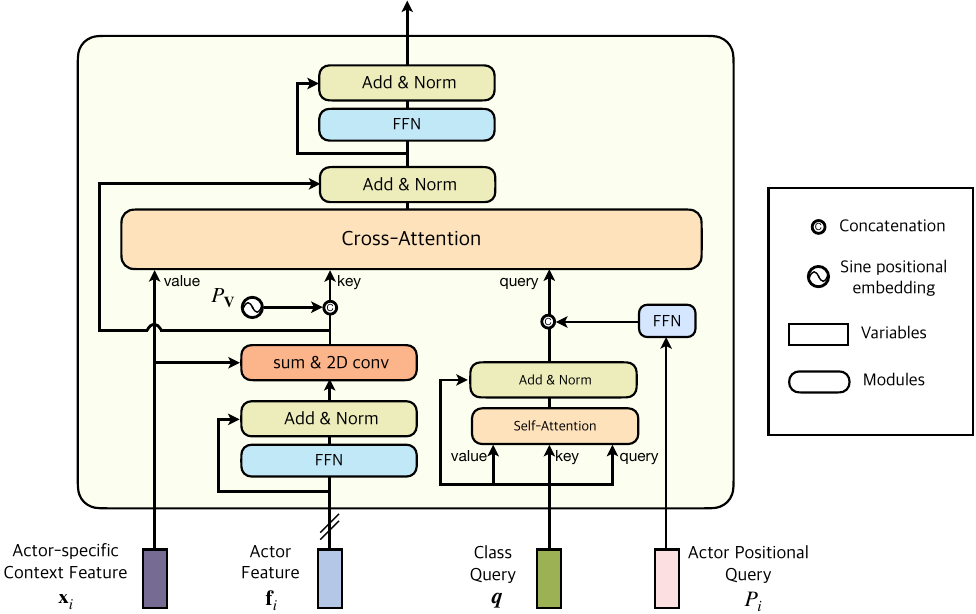}
     \caption{Structure of Classifying Decoder Layer (CDL)}
     \label{CDL_detailed}
     \vspace{-1mm}
\end{figure*}

We provide details of Classifying Decoder Layer (CDL), describing its miscellaneous operations that are omitted in the main paper.
Fig.~\ref{CDL_detailed} describes the process of CDL.
For brevity, we explain how CDL operates to infer the $i$-th actor's action class ($i\in [1, N_a]$).
Classifying decoder layer (CDL) takes as inputs the actor feature $\mathbf{f}_i \in \mathbb{R}^{D}$, the actor positional query ${P_i} \in \mathbb{R}^{D}$, and the actor-specific context features $\mathbf{x}_{i} \in \mathbb{R}^{T \times H \times W \times D}$, that are generated in LDL.
It additionally takes as its input the learnable embeddings $\boldsymbol{q} \in \mathbb{R}^{N_c \times D}$, which we denote as \emph{class queries}, to embed class-specific information.
To elaborate, CDL aims to construct a classification feature $\tilde{\boldsymbol{q}} \in \mathbb{R}^{N_c \times D}$ for each $i$-th actor, where $\tilde{\boldsymbol{q}}_c \in \mathbb{R}^{D}$ implies the feature used to detect $c$-th action performed by the $i$-th actor.


First, the actor feature passes through a standard feedforward network to be refined for classification.
However, it has been noted that optimization of the classification feature can sometimes interfere with localization~\cite{wu2020rethinking, song2020revisiting}.
Thus, the gradient flow is halted before this feature enters the classifying decoder layer.

We begin with creating features that incorporate the interaction between the $i$-th actor and the actor-specific context feature $\mathbf{x}_{i}$.
To this end, we spatially duplicate the actor feature $\mathbf{f}_i$ to the spatial extent of $\mathbf{x}_{i}$ and take the sum of these two features to obtain $(\mathbf{f} + \mathbf{x})_i \in \mathbb{R}^{H\times W \times D}$.
$(\mathbf{f} + \mathbf{x})_i$ is processed through multiple convolutional layers to embed local relationships between features in $(\mathbf{f} + \mathbf{x})_i$.
Let us denote the flattened resulting tensor $\mathbf{z}_i$ as follows:

\begin{equation}\label{aeqn:eq8}
{\mathbf{z}_i = \textnormal{Conv2D}\Bigl((\mathbf{f} + \mathbf{x})_i\Bigr) \in \mathbb{R}^{HW \times D}}.
\end{equation}

On the other hand, class queries are trained to encode class-specific information as the layers proceed.
Also, the self-attention mechanism between these queries enables them to exchange information among different classes and obtain enriched class-specific features that are aware of the occurrence of other classes.
To be specific, the class queries $\boldsymbol{q}$ are passed to the self-attention layer $f_{\text{SA}}^{\text{cls}}(q,k,v)$, serving themselves as the query, key, and value.
The output class queries $\boldsymbol{q}$ are then joined with $\mathbf{z}_i$ and $\mathbf{x}_i$ to construct a triplet of inputs for cross-attention layer $f_{\text{CA}}^{\textnormal{cls}}(q,k,v)$, in which the class queries collect information about each class from the features that represent the video context.

In the following cross-attention layer, $\boldsymbol{q}$ and $\mathbf{z}_i$ act as the query and key, respectively.
They are trained so that $\boldsymbol{q}_c$ and $\mathbf{z}_i[m]\ (m \in [1, HW])$ becomes semantically similar when the location $m$ is relevant to the $c$-th action class.
Specifically, the resulting attention map $\mathcal{A}_i \in \mathbb{R}^{HW \times N_c}$ signifies informative regions for each class and is multiplied with the value $\mathbf{x}_{i}$ to output class-specific features $\tilde{\boldsymbol{q}} \in \mathbb{R}^{N_c \times D}$.
However, we have discovered the $c$-th class query $\boldsymbol{q}_c$ often activates region where the $c$-th actions happen, but not the ones necessarily performed by the $i$-th actor.
Since the $c$-th class query learns to activate the region $\mathcal{A}_i[m]$ whenever the region $m$ is relevant to the $c$-th action, it occasionally confuses to which actor's action it should react.
To handle this actor confusion problem, we concatenated the global positional embedding $P_\mathbf{V}$ and the modulated actor positional queries $P_i$ to key and query, respectively.
Consequently, the resulting cross-attention $f_{\text{CA}}^{\textnormal{cls}}$ happens as follows:
\begin{equation}\label{aeqn:eq9}
    {\tilde{\boldsymbol{q}} \leftarrow f_{\text{CA}}^{\textnormal{cls}}\Bigl(\textsl{q = }\bigl[\boldsymbol{q}, P_i \bigr],}\ 
    {\textsl{k = }\bigl[\mathbf{z}_{i}, P_\mathbf{V}\bigr],}\ 
    {\textsl{v = }\mathbf{x}_{i} \Bigr) }.
\end{equation}
As the LDL's output feature of Eq.~\ref{aeqn:eq6} is subject to localization, the spatial parts $P_\mathbf{V}$ and $P_i$ are trained to obtain the positional information of the $i$-th actor.
Hence, concatenating these positional information provides subtle cues about the actor's position to class queries and adds actor-specificity to the output $\tilde{\boldsymbol{q}}$.

The output $\tilde{\boldsymbol{q}} \in \mathbb{R}^{N_c \times D}$ of the cross-attention passes the ordinary feedforward network and is subsequently passed to the next layer to serve as class queries again.
To derive the probability for each class, $\tilde{\boldsymbol{q}}$ from the last layer is mean pooled across the channel dimension and then processed through a sigmoid layer.
Thus, the final output $\tilde{\boldsymbol{q}}\in [0,1]^{N_c}$ becomes the classification score for the $i$-th actor.

\section{Details of the Hungarian matching process}\label{appendix:Hungarian_matching}

For the Hungarian matching between the $i$-th element $Y_i = \bigl(\mathbf{B}_{i}, \mathbf{C}_{i}\bigr) $ in the padded ground-truth labels 
and the $j$-th element $\hat{Y}_j = \bigl(\hat{\mathbf{B}}_j, \hat{\mathbf{C}}_j\bigr) $ in the model predictions, we consider three matching costs, 
$\mathcal{H}_{i,j}^{\textnormal{box}}$, $\mathcal{H}_{i,j}^{\textnormal{giou}}$, and $\mathcal{H}_{i,j}^{\textnormal{class}}$, which are defined as follows:
\begin{equation}\label{hungarian costs}
    \begin{split}
        \mathcal{H}_{i,j}^{\textnormal{box}} &= \lVert{\mathbf{B}_i - \hat{\mathbf{B}}_j}\rVert_1, \\
        \mathcal{H}_{i,j}^{\textnormal{GIoU}} &=
        -{\text{GIoU}}(\mathbf{B}_i, \hat{\mathbf{B}}_j\bigr), \\
        \mathcal{H}_{i,j}^{\textnormal{class}} &=
        {\text{BCELoss}}(\mathbf{C}_i, \hat{\mathbf{C}}_j), \\
    \end{split}
\end{equation}
where ${\text{GIoU}}(\cdot, \cdot)$ is a generalized IoU~\cite{rezatofighi2019generalized} between boxes and ${\text{BCELoss}}(\cdot, \cdot)$ is a binary cross entropy loss between two multi-hot vectors.
Then, the Hungarian algorithm aims to find the optimal assignment ${\hat{\omega} = \argmin_{\omega \in \boldsymbol{\Omega}_{N_a}} \sum_{i=1}^{N_a}\bigl( \eta_{\textnormal{box}}\mathcal{H}_{i,\omega(i)}^{\textnormal{box}}} \\ + \eta_{\textnormal{GIoU}}\mathcal{H}_{i,\omega(i)}^{\textnormal{GIoU}}$

${+ \eta_{\textnormal{class}}\mathcal{H}_{i,\omega(i)}^{\textnormal{class}}} \bigr)$,
where $\eta_{\textnormal{box}}, \eta_{\textnormal{box}},$ and $\eta_{\textnormal{box}}$ are coefficients that balance the matching cost for each term.
In the case of AVA, we figure that setting $\mathcal{H}_{i,j}^{\textnormal{class}}$ as binary cross entropy loss leads to a slow convergence, which is potentially due to its multi-label property.
Thus, we set $\mathcal{H}_{i,j}^{\textnormal{class}}$ as $-\hat{\boldsymbol{p}}_{j}$, since the confidence $\hat{\boldsymbol{p}}_{j}$ is learned to have a higher value if $j$-th element is matched with a ground-truth label.
Note that such matching method inherits the way from TubeR~\cite{zhao2022tuber}.

\section{Comparison with latest VAD models}\label{appendix:model_comparison}

We provide a comparison analysis between ours and latest VAD models, TubeR~\cite{zhao2022tuber}, STMixer~\cite{wu2023stmixer}, and EVAD~\cite{chen2023efficient}.

\noindent \textbf{Ours vs TubeR.}
TubeR is similar to ours in that it is established on DETR~\cite{carion2020end} architecture along with tube-shaped queries.
Furthermore, it outputs multiple frames with a single feed-forward as well, showing greater efficiency compared to other methods.
However, the encoder stage of TubeR consumes substantial memory since it performs self-attention over spatio-temporal features.

\noindent \textbf{Ours vs STMixer.}
Our model and STMixer are similar in that they both utilize multi-scale feature maps to capture fine details of an actor's action.
While ours has the deformable encoding process (Sec.~\ref{appendix:3DDefEnc}), STMixer does not explicitly have a distinguishable encoding module.
However, it has an Adaptive Feature Sampling module, which resembles the deformable encoding process: it also determines which feature points to sample from the multi-scale feature maps with learnable parameters. 

\noindent \textbf{Ours vs EVAD.}
EVAD, in contrast, does not employ multi-scale feature maps.
Instead, EVAD takes advantage of encoding the video features in a dense manner: although it prunes the features and makes the self-attention process less heavy, EVAD still exhaustively computes self-similarities between each feature vector and obtains richer encoded features.
In addition, EVAD uses the final localization output to obtain classification features, which differs from ours in which the localization and classification features evolve together in the decoder module.

\noindent \textbf{Comparison in a module-level.}
\begin{table}[t]
\caption{Comparison with other models~\cite{chen2023efficient, zhao2022tuber} in terms of module components. $^{\ddag}$ is to indicate that its weight is pretrained on COCO~\cite{Mscoco}. CSN-152~\cite{tran2019video} backbone is used.}
\centering
\begin{tabular}{lllc}
\toprule
Method \qquad\quad & {Encoder} & {Decoder } & mAP \\ 
\hline
\\[-1em]
- & Transformer & DETR~\cite{carion2020end} & 28.6 \\
- & Transformer & LDL & 29.1 \\
- & Transformer & LDL + CDL & 31.4 \\
TubeR~\cite{zhao2022tuber} & Transformer{$^\ddag$} + LSTR Decoder~\cite{xu2021long} \qquad  & DETR{$^\ddag$} + Transformer & 31.1 \\
EVAD~\cite{chen2023efficient} & Transformer + KTP~\cite{chen2023efficient} & FPN~\cite{lin2017_fpn} + Transformer & N/A \\
- & 3D Deformable Transformer \ & LDL & 31.3 \\
STMixer~\cite{wu2023stmixer} & Adaptive Feature Sampling~\cite{wu2023stmixer} & Adaptive Feature Mixing~\cite{wu2023stmixer} & 32.8 \\
Ours & 3D Deformable Transformer \ & LDL + CDL & 33.5 \\
\bottomrule
\end{tabular}
\label{module_ablations}
\end{table}
We provide a module-level comparison Table~\ref{module_ablations} for a thorough comparison with ours and previous methods.
Details for each model are as follows:
\begin{itemize}
    \item TubeR~\cite{zhao2022tuber}
    \begin{itemize}
        \item uses traditional transformer encoder, but also utilizes Long Short-Term Transformer~\cite{xu2021long} for temporal aggregation.
        \item employs DETR~\cite{carion2020end} weights pretrained on COCO~\cite{Mscoco}.
    \end{itemize}
    \item STMixer~\cite{wu2023stmixer}
    \begin{itemize}
        \item has no explicit encoding module, but based on the roles of their modules, we categorize Adaptive Feature Sampling and Adaptive Feature Mixing modules as its encoder and decoder, respectively.
        \item introduces Adaptive Feature Sampling, which resembles the mechanism of Deformable-DETR~\cite{zhu2020deformable} in that features are encoded by sampling from the feature map.
        \item decodes the sampled feature with Adaptive Feature Mixing, where the queries generate parameters of the modules through which they pass, similar to Sparse R-CNN~\cite{sun2021sparse}.
    \end{itemize}   
    \item EVAD~\cite{chen2023efficient}
    \begin{itemize}
        \item tokenizes video frames and prunes the tokens inside its encoder module.
        The tokens are mainly dropped besides the keyframe of the clip, hence it is named Keyframe-centric Token Pruning (KTP).
        \item employs Feature Pyramid Network~\cite{lin2017_fpn} for its localization module.
        \item does not report its performance on CSN-152, so it is marked `N/A'.
    \end{itemize}        
\end{itemize}

\begin{table}[t]
\caption{Comparison with other models~\cite{chen2023efficient, zhao2022tuber} in terms of computational complexity.}
\centering
\begin{tabular}{lllcc}
\toprule
Method \qquad\quad & {Encoder} & {Decoder} & $N_a$ & K \\ 
\hline
\\[-1em]
TubeR~\cite{zhao2022tuber} & $\mathcal{O}((HW)^2)$ \qquad  & $\mathcal{O}((HW)^2 + HW \times N_a)$ & 15 & - \\
EVAD~\cite{chen2023efficient} & $\mathcal{O}((\rho \times THW)^2)$ & $\mathcal{O}((\rho^3 \times THW + N_a)^2)$ & 100 & -\\
STMixer~\cite{wu2023stmixer} &$\mathcal{O}(N_a \times K)$ & $\mathcal{O}(K\times D + N_a \times D)$ & 100 & 32 \\
Ours & $\mathcal{O}(HW \times K)$ & $\mathcal{O}(HW \times N_a + N_a \times N_c)$ & 15 & 4\\
\bottomrule
\end{tabular}
\label{efficiency_analysis}
\end{table}
\subsubsection{Efficiency analysis.}
We present computational complexities of the latest models in Table~\ref{efficiency_analysis}.
The major bottleneck of TubeR~\cite{zhao2022tuber} and EVAD~\cite{chen2023efficient} comes from exhaustive self-attention between $HW$ feature vectors.
Specifically, regarding that the flattened spatial size $HW$, which typically falls in between $300$ and $400$, is larger than other variables such as $N_a$ or $K$, this operation creates a significant computational burden on both encoder and decoder architecture.
On the other hand, STMixer~\cite{wu2023stmixer} and ours resolve such computational burden of the encoder by sparsely collecting features from multi-level feature maps.
However, STMixer generates adaptive parameters in its decoder architecture, and it results in complexities involving the channel dimension $D$, which typically is set to $256$.
Ours does not involve $D$, but still performs dense cross-attention using $HW$ feature vectors.
Yet, ours acquires greater efficiency by occupying much smaller number of actor candidates $N_a$.

In fact, the reason for our model being able to surpass other methods with comparably smaller $N_a$ is benefited from the utilization of class queries.
EVAD and STMixer extensively increase their number of actor candidates to obtain a pool of sufficiently diverse classification maps, which eventually improves their performance since such diversification increases a chance of capturing overlooked context needed to classify the action of the target actor.
It is a crucial strategy for these models: because they expect the classification attention map for each actor to capture shared semantics of every action class, clues to identify some action classes (\eg, clues that are distant from the actor) are easily missed when $N_a$ is small.
In contrast, our model can generate more precise attention maps for each actor and does not need to rely on large number of actor candidates.
We utilize this saved memory to enable our model to process multiple frames with a single feedforward, maximizing the efficiency created from using class queries.

\section{Experimental setup}\label{appendix:experimental_setup}
\subsubsection{Network configurations.}
We set the channel dimension $D$ to 256, number of feature levels $L$ to 4, number of sampling points in the encoder $K$ to 8.
Otherwise, the model hyperparameters differ across three datasets, so we specify them in Table~\ref{implementation details}.
\subsubsection{Training details.}
We use AdamW~\cite{loshchilov2017decoupled} optimizer and linearly warm up the learning rate in earlier epochs.
The step scheduler is applied, so the learning rate decays by 0.1 every step milestone.
Following prior work~\cite{chen2023efficient, wu2023stmixer}, standard data augmentation techniques such as color jittering, random horizontal flipping, and PCA jittering~\cite{krizhevsky2012imagenet} are applied.
We present further training configurations in ~\ref{training details}.

Additionally, we have observed that the model struggles to converge from scratch at once.
Hence, we follow the training practice introduced in TubeR~\cite{zhao2022tuber}.
TubeR utilizes DETR~\cite{carion2020end} weights pretrained on COCO~\cite{Mscoco} for its transformer module and fine-tune these weights on AVA before the actual training.
The fine-tuned transformer weights are then used for the final training: the transformer module of the model is trained from the pretrained weights, while the remaining modules are trained from scratch.
Similarly, we first train the model from scratch, possibly on smaller backbones for ease of training, and then use the transformer weights acquired from the first stage for the final training.

\setlength{\floatsep}{8pt plus 0pt minus 10pt}
\begin{table}[t!]
\centering
\caption{Network configurations for each dataset.}
\begin{tabular}{lcccc}
\toprule
Hyper- & \multirow{2}{*}{Description} & \multirow{2}{*}{AVA} & \multirow{2}{*}{JHMDB} & \multirow{2}{*}{UCF} \\
parameter & & & & \\ [0.5em]
\hline
\\[-1em]
\multirow{2}{*}{$W_0$}     & resized resolution & \multirow{2}{*}{256} & \multirow{2}{*}{288} & \multirow{2}{*}{256}\\
\quad   & (length of the shorter side) & & & \\
$T_0$   & clip length & 32 or 16 & 40 & 32 \\
$N_a$   & number of actor candidates & 15 & 5 & 15 \\
$N_{\textnormal{enc}}$ & number of encoder layers & 6 & 3 & 3 \\
$N_{\textnormal{dec}}$     & number of decoder layers & 6 & 3 & 3 \\
\multirow{2}{*}{$N_c$}     & number of class queries & \multirow{2}{*}{80} & \multirow{2}{*}{21} & \multirow{2}{*}{24}\\
\quad   & (number of classes) & & & \\

\bottomrule
\end{tabular}

\label{implementation details}
\end{table}
\setlength{\floatsep}{12pt plus 0pt minus 10pt}
\begin{table}[t!]
\centering
\caption{Training configurations for each dataset.}
\begin{tabular}{lcccc}
\toprule
\multicolumn{2}{l}{Training configurations} & {AVA} &{JHMDB} &{UCF} \\ [0.5em]
\hline
\\[-1em]
\multicolumn{2}{c}{learning rate} & 1e-4 & 2e-4 & 2e-4 \\
\multicolumn{2}{c}{learning rate milestone} & [8,11] & [50] & [12] \\
\multicolumn{2}{c}{warmup start learning rate} & 1e-5 & 2e-6 & 2e-6 \\
\multicolumn{2}{c}{weight decay} & 1e-4 & 2e-4 & 2e-4 \\
\multicolumn{2}{c}{epochs} & 12 & 100 & 14 \\
\multicolumn{2}{c}{warmup epochs} & 3 & 10 & 3 \\
$ $     & loss coefficient for class ($\lambda_{\textnormal{class}}$)& 10 & 4 & 4 \\
$ $     & loss coefficient for box ($\lambda_{\textnormal{box}}$)& 5 & 5 & 5 \\
$ $     & loss coefficient for GIoU ($\lambda_{\textnormal{giou}}$)& 2 & 2 & 2 \\
$ $     & loss coefficient for confidence ($\lambda_{\textnormal{conf}}$)& 1 & 6 & 3 \\

\bottomrule
\end{tabular}

\label{training details}
\end{table}

\section{Additional Ablation Experiments}\label{appendix:additional_ablations}

\textbf{Effectiveness of box modulation.} We adopt the box modulator function~\cite{liu2022dab} to effectively utilize the positional prior of the actor box.
Specifically, it enhances the modulation process by involving the height and width information of the actor box.
Hence, the actor positional queries can also be applied without the box modulator, thereby encoding the positional information with positional coordinates without its height and width prior.
It turns out that such modulation improves the performance by 0.8\% in AVA and by 0.2\% in UCF, as demonstrated in Table~\ref{box_modulator}.

\noindent \textbf{Effectiveness of class query self-attention.} 
We allow class queries to attend to each other before they enter the transformer module.
The motivation behind this process is to let the model consider the relationship between different classes.
For example, while the class \textsl{`stand'} and \textsl{`sit'} can never co-occur at the same time, \textsl{`eat'} and \textsl{`carry/hold (an object)'} frequently happen simultaneously.
The effect of the self-attention between the class queries is shown in Table~\ref{cqsa}.

\setlength{\floatsep}{8pt plus 0pt minus 10pt}
\begin{figure}[t!]
    \begin{minipage}[t]{0.55\textwidth}
        \centering
        \captionof{table}{Ablation experiments on utilizing the box modulator~\cite{liu2022dab}.}
        \vspace{1mm}
        \begin{tabular}{lcc}
    \toprule
    {Method}  & AVA & UCF \\ 
    \hline
    \\[-1em]
    w/ modulated actor positional queries & 33.5 & 85.9 \\
    w/ ordinary actor positional queries & 32.7 & 84.8 \\
    \bottomrule
\end{tabular}\label{abl:box_modulator}\label{box_modulator}
    \end{minipage}
    \hspace{0.9em}
    \begin{minipage}[t]{0.38\textwidth}
        \centering
        \captionof{table}{The effect of self-attention between class queries.}
        \vspace{1mm}
        \resizebox{\textwidth}{!}{
\begin{tabular}{lcc}
    \toprule
    Method & AVA  & UCF \\ 
    \hline
    \\[-1em]
    w/ class query SA & 33.5  & 85.9 \\
    w/o class query SA & 31.6  & 83.2 \\ 
    \bottomrule
\end{tabular}
}\label{query_sa}\label{cqsa}
    \end{minipage}
\vspace{4mm}

\end{figure}
\begin{figure}[t!]
     \centering
     \includegraphics[width=\textwidth]{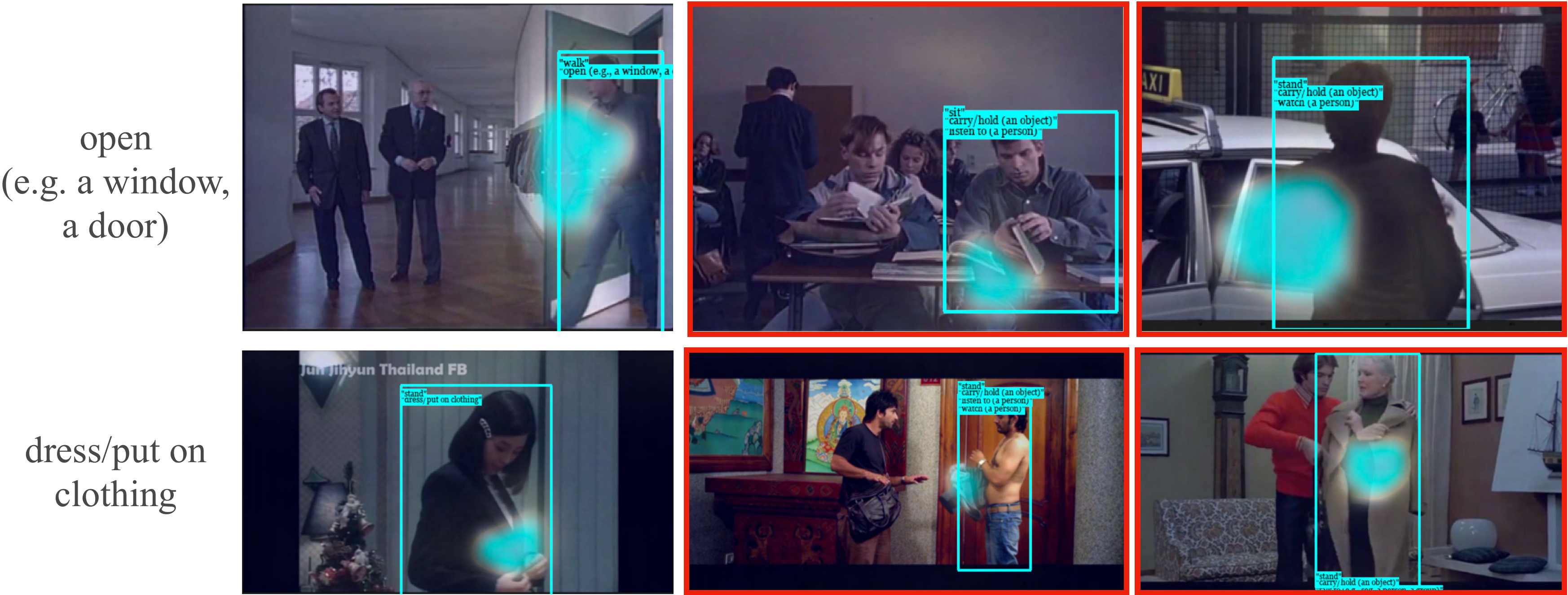}
     \vspace{-5mm}
     \caption{Failure cases are marked with red boundaries.}
     \label{fig:failure_cases}
\end{figure}
\section{More qualitative results}\label{appendix:more_qualitative}

\noindent \textbf{Failure cases.}
We observe that our model occasionally uses spurious biases, especially in cases where a class covers various scenarios: \eg, the class \textsl{`open'} is labeled not only when an actor opens an ordinary door but also a car door or a book.
The same happens for \textsl{`dress/put on clothing'}, in which there are many ways to dress up various clothing (Fig.~\ref{fig:failure_cases}).

\begin{figure*}[t!]
     \centering
     \includegraphics[width=\textwidth]{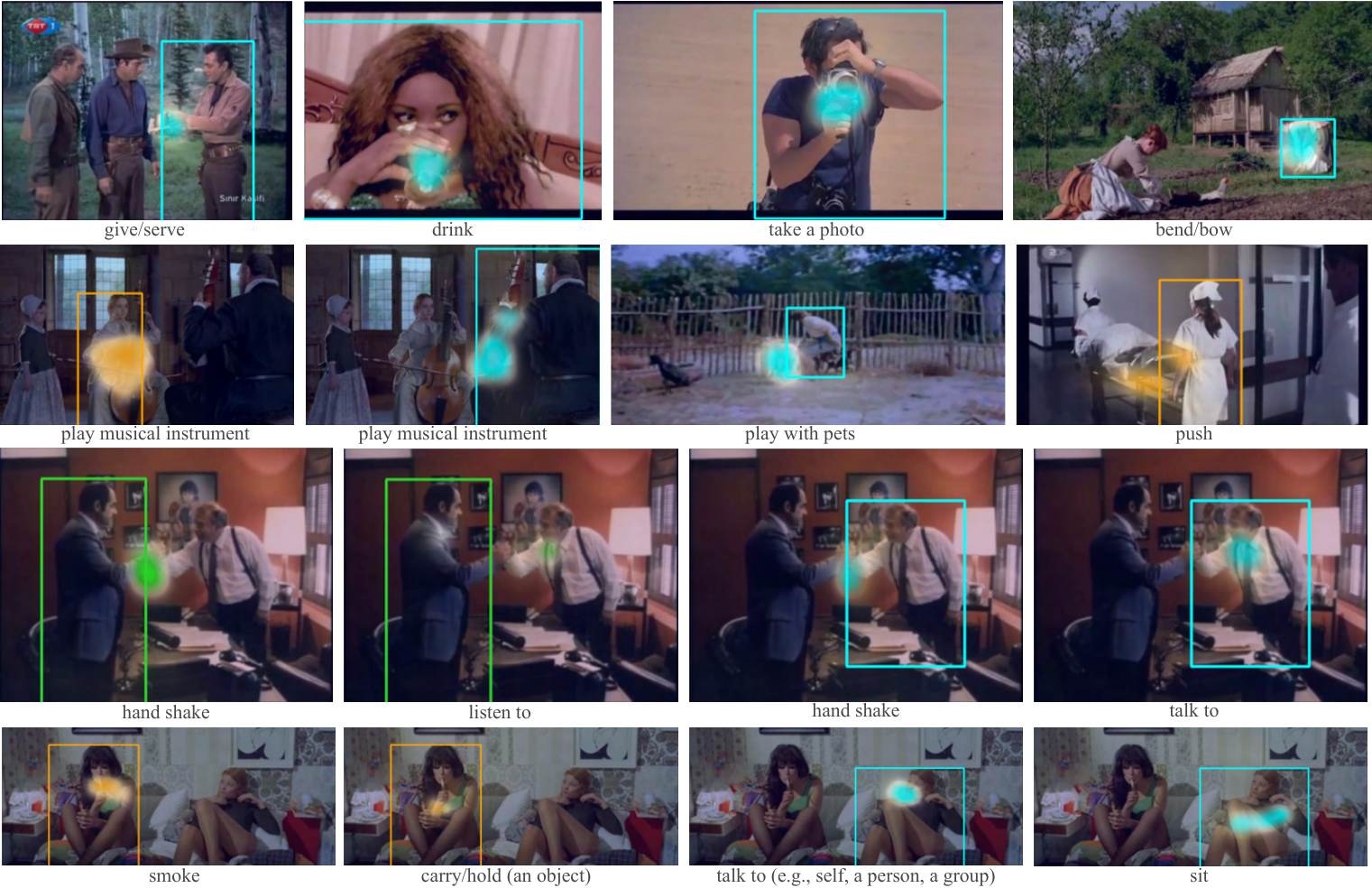}
     \caption{Classification attention map on AVA.}
     \label{ava_vis}
\end{figure*}

\begin{figure*}[t!]
     \centering
     \includegraphics[width=\textwidth]{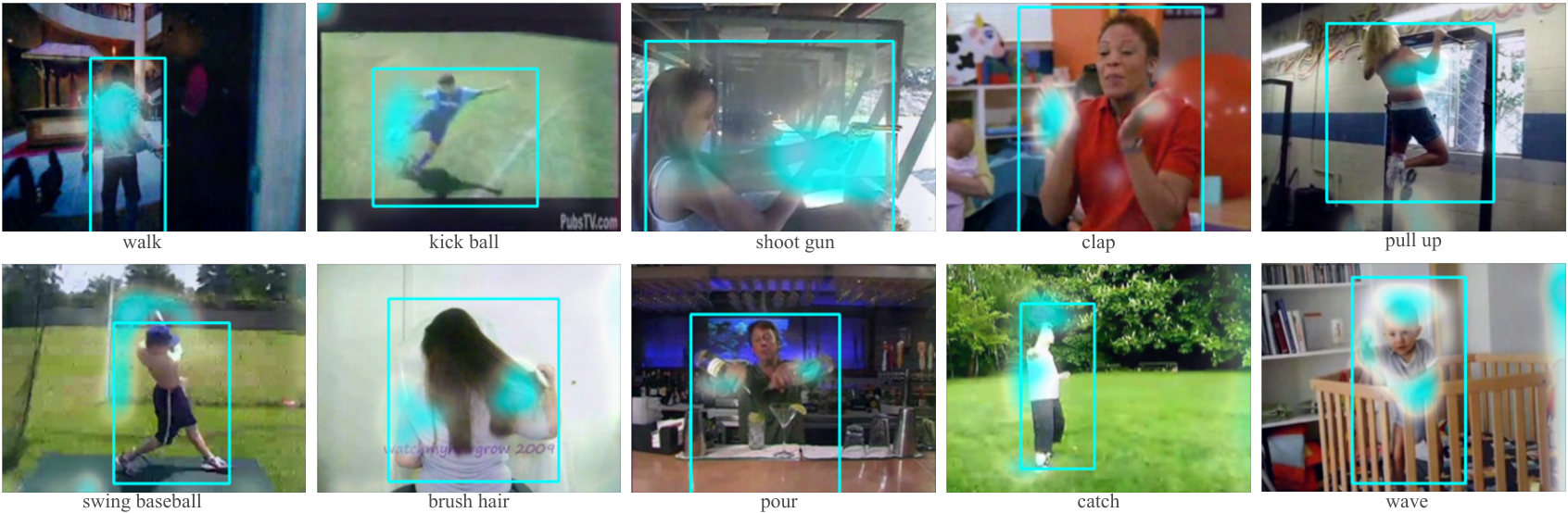}
     \caption{Classification attention map on JHMDB51-21.}
     \label{jhmdb_vis}
\end{figure*}

\begin{figure*}[t!]
     \centering
     \includegraphics[width=\textwidth]{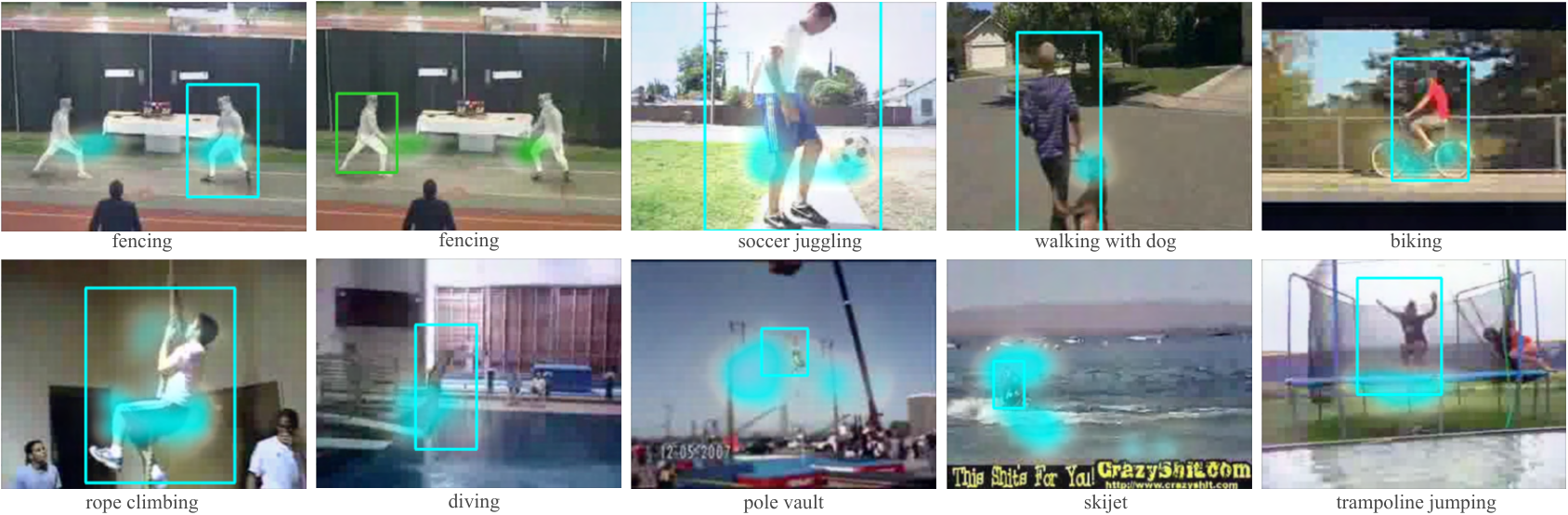}
     \caption{Classification attention map for UCF101-24.}
     \label{ucf_vis}
\end{figure*}

\noindent \textbf{Classification attention map on other datasets.}
Fig.~\ref{ava_vis},~\ref{jhmdb_vis}, and~\ref{ucf_vis} illustrate the classification attention map visualization results from AVA, JHMDB51-21, and UCF101-24, respectively.
Our model concentrates on areas that are crucial for classification and they are not necessarily on the actor's body.
For example, the model sees an animal (Fig.~\ref{ava_vis}: \textsl{play with pets}, Fig.~\ref{ucf_vis}: \textsl{walking with dog}) or an object (Fig.~\ref{ava_vis}: \textsl{push}, Fig.~\ref{jhmdb_vis}: \textsl{pour}, Fig.~\ref{ucf_vis}: \textsl{pole vault}, \textsl{trampoline jumping}, \textsl{biking}) which is an essential clue to understand the action happening in the video clip.

\section{Class-wise mAP comparison}\label{appendix:classwise_map}

\begin{figure*}[t!]
     \centering\
     \includegraphics[width=\textwidth]{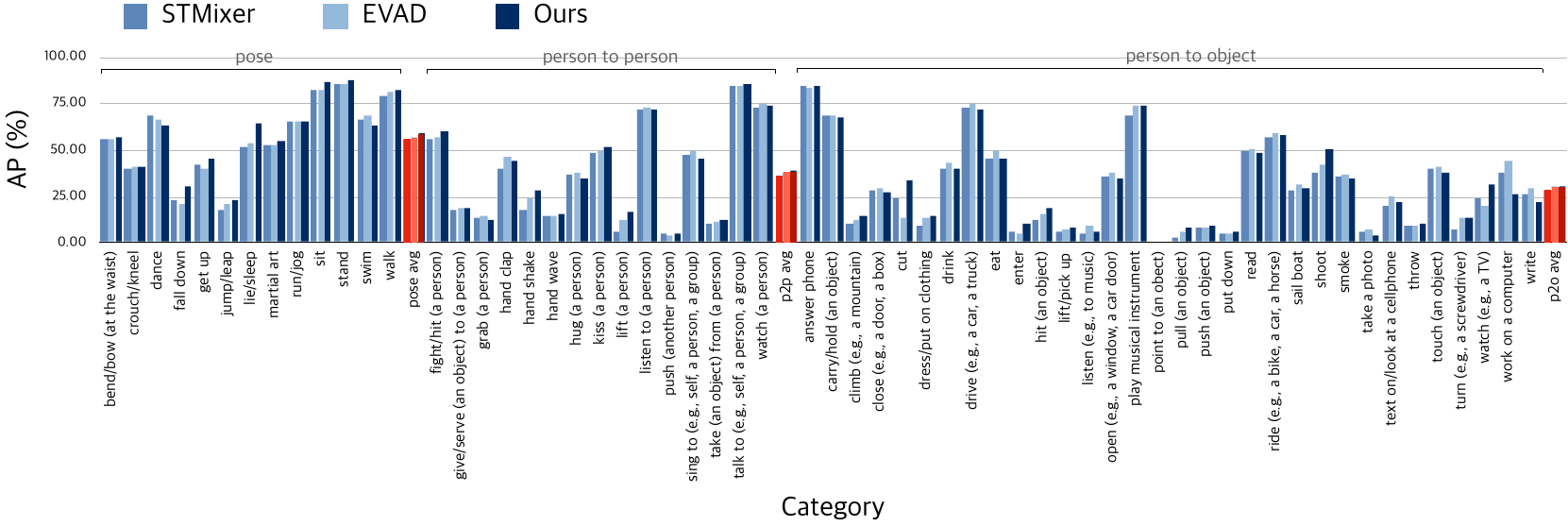}
     \caption{Class-wise comparison among latest VAD models, STMixer~\cite{wu2023stmixer} and EVAD~\cite{chen2023efficient}. Red bars are mean value that aggregates for each category type: pose, person to person, and person to object.}
     \label{classwise comparison}
\end{figure*}

We provide the comparison chart (Fig.~\ref{classwise comparison}) by the class labels in AVA.
As the comparison group, we choose EVAD~\cite{chen2023efficient} and STMixer~\cite{wu2023stmixer} trained on ViT-B~\cite{wang2023videomae}.
Out of total 60 class labels, our model demonstrates the best performance on 33 labels, while EVAD and STMixer show their superiorities on 24 and 3 labels, respectively.


\setlength{\textfloatsep}{0pt plus 0pt minus 5pt}
\setlength{\textfloatsep}{5pt plus 0pt minus 5pt}

\setlength{\textfloatsep}{0pt plus 0pt minus 5pt}
\setlength{\textfloatsep}{5pt plus 0pt minus 5pt}

\setlength{\textfloatsep}{0pt plus 0pt minus 5pt}
\setlength{\textfloatsep}{5pt plus 0pt minus 5pt}

\setlength{\textfloatsep}{5pt plus 0pt minus 5pt}
\setlength{\textfloatsep}{30pt plus 5pt minus 0pt}

\clearpage



\section*{Acknowledgements}

This work was partly supported by the NAVER Cloud Corporation and the NRF grant and IITP grant funded by Ministry of Science and ICT, Korea 
(NRF-2021R1A2C3012728,
RS-2022-II220926,
RS-2022-II220290,
RS-2019-II191906).
\bibliographystyle{splncs04}
\bibliography{references}

\begin{thebibliography}{10}
\providecommand{\url}[1]{\texttt{#1}}
\providecommand{\urlprefix}{URL }
\providecommand{\doi}[1]{https://doi.org/#1}

\bibitem{carion2020end}
Carion, N., Massa, F., Synnaeve, G., Usunier, N., Kirillov, A., Zagoruyko, S.: End-to-end object detection with transformers. In: ECCV (2020)

\bibitem{chen2023efficient}
Chen, L., Tong, Z., Song, Y., Wu, G., Wang, L.: Efficient video action detection with token dropout and context refinement. arXiv preprint arXiv:2304.08451  (2023)

\bibitem{chen2021watch}
Chen, S., Sun, P., Xie, E., Ge, C., Wu, J., Ma, L., Shen, J., Luo, P.: Watch only once: An end-to-end video action detection framework. In: Proceedings of the IEEE/CVF International Conference on Computer Vision. pp. 8178--8187 (2021)

\bibitem{dosovitskiy2021an}
Dosovitskiy, A., Beyer, L., Kolesnikov, A., Weissenborn, D., Zhai, X., Unterthiner, T., Dehghani, M., Minderer, M., Heigold, G., Gelly, S., Uszkoreit, J., Houlsby, N.: An image is worth 16x16 words: Transformers for image recognition at scale. In: ICLR (2021)

\bibitem{feichtenhofer2020x3d}
Feichtenhofer, C.: X3d: Expanding architectures for efficient video recognition. In: Proceedings of the IEEE/CVF conference on computer vision and pattern recognition. pp. 203--213 (2020)

\bibitem{feichtenhofer2019slowfast}
Feichtenhofer, C., Fan, H., Malik, J., He, K.: Slowfast networks for video recognition. In: Proceedings of the IEEE/CVF international conference on computer vision. pp. 6202--6211 (2019)

\bibitem{ghadiyaram2019large}
Ghadiyaram, D., Tran, D., Mahajan, D.: Large-scale weakly-supervised pre-training for video action recognition. In: Proceedings of the IEEE/CVF conference on computer vision and pattern recognition. pp. 12046--12055 (2019)

\bibitem{gu2018ava}
Gu, C., Sun, C., Ross, D.A., Vondrick, C., Pantofaru, C., Li, Y., Vijayanarasimhan, S., Toderici, G., Ricco, S., Sukthankar, R., et~al.: Ava: A video dataset of spatio-temporally localized atomic visual actions. In: Proceedings of the IEEE conference on computer vision and pattern recognition. pp. 6047--6056 (2018)

\bibitem{guo2020augfpn}
Guo, C., Fan, B., Zhang, Q., Xiang, S., Pan, C.: Augfpn: Improving multi-scale feature learning for object detection. In: Proceedings of the IEEE/CVF conference on computer vision and pattern recognition. pp. 12595--12604 (2020)

\bibitem{herzig2022object}
Herzig, R., Ben-Avraham, E., Mangalam, K., Bar, A., Chechik, G., Rohrbach, A., Darrell, T., Globerson, A.: Object-region video transformers. In: Proceedings of the IEEE/CVF Conference on Computer Vision and Pattern Recognition. pp. 3148--3159 (2022)

\bibitem{jhuang2013towards}
Jhuang, H., Gall, J., Zuffi, S., Schmid, C., Black, M.J.: Towards understanding action recognition. In: Proceedings of the IEEE international conference on computer vision. pp. 3192--3199 (2013)

\bibitem{josmy2022holistic}
Josmy~Faure, G., Chen, M.H., Lai, S.H.: Holistic interaction transformer network for action detection. arXiv e-prints pp. arXiv--2210 (2022)

\bibitem{kay2017kinetics}
Kay, W., Carreira, J., Simonyan, K., Zhang, B., Hillier, C., Vijayanarasimhan, S., Viola, F., Green, T., Back, T., Natsev, P., et~al.: The kinetics human action video dataset. arXiv preprint arXiv:1705.06950  (2017)

\bibitem{kopuklu2019you}
K{\"o}p{\"u}kl{\"u}, O., Wei, X., Rigoll, G.: You only watch once: A unified cnn architecture for real-time spatiotemporal action localization. arXiv preprint arXiv:1911.06644  (2019)

\bibitem{krizhevsky2012imagenet}
Krizhevsky, A., Sutskever, I., Hinton, G.E.: Imagenet classification with deep convolutional neural networks. Advances in neural information processing systems  \textbf{25} (2012)

\bibitem{kuhn1955hungarian}
Kuhn, H.W.: The hungarian method for the assignment problem. Naval research logistics quarterly  \textbf{2}(1-2),  83--97 (1955)

\bibitem{li2022dn}
Li, F., Zhang, H., Liu, S., Guo, J., Ni, L.M., Zhang, L.: Dn-detr: Accelerate detr training by introducing query denoising. In: Proceedings of the IEEE/CVF Conference on Computer Vision and Pattern Recognition. pp. 13619--13627 (2022)

\bibitem{li2020actions}
Li, Y., Wang, Z., Wang, L., Wu, G.: Actions as moving points. In: Computer Vision--ECCV 2020: 16th European Conference, Glasgow, UK, August 23--28, 2020, Proceedings, Part XVI 16. pp. 68--84. Springer (2020)

\bibitem{lin2017_fpn}
Lin, T.Y., Doll\'{a}r, P., Girshick, R., He, K., Hariharan, B., Belongie, S.: Feature pyramid networks for object detection. In: CVPR (2017)

\bibitem{lin2017focal}
Lin, T.Y., Goyal, P., Girshick, R., He, K., Doll{\'a}r, P.: Focal loss for dense object detection. In: Proceedings of the IEEE international conference on computer vision. pp. 2980--2988 (2017)

\bibitem{Mscoco}
Lin, T.Y., Maire, M., Belongie, S., Hays, J., Perona, P., Ramanan, D., Doll{\'a}r, P., Zitnick, C.L.: {Microsoft COCO:} common objects in context. In: ECCV (2014)

\bibitem{liu2022dab}
Liu, S., Li, F., Zhang, H., Yang, X., Qi, X., Su, H., Zhu, J., Zhang, L.: Dab-detr: Dynamic anchor boxes are better queries for detr. In: ICLR (2022)

\bibitem{loshchilov2017decoupled}
Loshchilov, I., Hutter, F.: Decoupled weight decay regularization. ICLR  (2019)

\bibitem{meng2021conditional}
Meng, D., Chen, X., Fan, Z., Zeng, G., Li, H., Yuan, Y., Sun, L., Wang, J.: Conditional detr for fast training convergence. In: Proceedings of the IEEE/CVF International Conference on Computer Vision. pp. 3651--3660 (2021)

\bibitem{pan2021actor}
Pan, J., Chen, S., Shou, M.Z., Liu, Y., Shao, J., Li, H.: Actor-context-actor relation network for spatio-temporal action localization. In: Proceedings of the IEEE/CVF Conference on Computer Vision and Pattern Recognition. pp. 464--474 (2021)

\bibitem{pang2020multi}
Pang, Y., Zhao, X., Zhang, L., Lu, H.: Multi-scale interactive network for salient object detection. In: Proceedings of the IEEE/CVF conference on computer vision and pattern recognition. pp. 9413--9422 (2020)

\bibitem{peng2016multi}
Peng, X., Schmid, C.: Multi-region two-stream r-cnn for action detection. In: Computer Vision--ECCV 2016: 14th European Conference, Amsterdam, The Netherlands, October 11--14, 2016, Proceedings, Part IV 14. pp. 744--759. Springer (2016)

\bibitem{rezatofighi2019generalized}
Rezatofighi, H., Tsoi, N., Gwak, J., Sadeghian, A., Reid, I., Savarese, S.: Generalized intersection over union: A metric and a loss for bounding box regression. In: Proceedings of the IEEE/CVF conference on computer vision and pattern recognition. pp. 658--666 (2019)

\bibitem{saha2016deep}
Saha, S., Singh, G., Sapienza, M., Torr, P.H., Cuzzolin, F.: Deep learning for detecting multiple space-time action tubes in videos. arXiv preprint arXiv:1608.01529  (2016)

\bibitem{singh2017online}
Singh, G., Saha, S., Sapienza, M., Torr, P.H., Cuzzolin, F.: Online real-time multiple spatiotemporal action localisation and prediction. In: Proceedings of the IEEE International Conference on Computer Vision. pp. 3637--3646 (2017)

\bibitem{song2020revisiting}
Song, G., Liu, Y., Wang, X.: Revisiting the sibling head in object detector. In: Proceedings of the IEEE/CVF conference on computer vision and pattern recognition. pp. 11563--11572 (2020)

\bibitem{soomro2012ucf101}
Soomro, K., Zamir, A.R., Shah, M.: Ucf101: A dataset of 101 human actions classes from videos in the wild. arXiv preprint arXiv:1212.0402  (2012)

\bibitem{sui2023simple}
Sui, L., Zhang, C.L., Gu, L., Han, F.: A simple and efficient pipeline to build an end-to-end spatial-temporal action detector. In: Proceedings of the IEEE/CVF Winter Conference on Applications of Computer Vision. pp. 5999--6008 (2023)

\bibitem{sun2018actor}
Sun, C., Shrivastava, A., Vondrick, C., Murphy, K., Sukthankar, R., Schmid, C.: Actor-centric relation network. In: Proceedings of the European Conference on Computer Vision (ECCV). pp. 318--334 (2018)

\bibitem{sun2021sparse}
Sun, P., Zhang, R., Jiang, Y., Kong, T., Xu, C., Zhan, W., Tomizuka, M., Li, L., Yuan, Z., Wang, C., et~al.: Sparse r-cnn: End-to-end object detection with learnable proposals. In: Proceedings of the IEEE/CVF conference on computer vision and pattern recognition. pp. 14454--14463 (2021)

\bibitem{tang2020asynchronous}
Tang, J., Xia, J., Mu, X., Pang, B., Lu, C.: Asynchronous interaction aggregation for action detection. In: Computer Vision--ECCV 2020: 16th European Conference, Glasgow, UK, August 23--28, 2020, Proceedings, Part XV 16. pp. 71--87. Springer (2020)

\bibitem{tong2022videomae}
Tong, Z., Song, Y., Wang, J., Wang, L.: Videomae: Masked autoencoders are data-efficient learners for self-supervised video pre-training. Advances in neural information processing systems  \textbf{35},  10078--10093 (2022)

\bibitem{tran2019video}
Tran, D., Wang, H., Torresani, L., Feiszli, M.: Video classification with channel-separated convolutional networks. In: Proceedings of the IEEE/CVF international conference on computer vision. pp. 5552--5561 (2019)

\bibitem{vaswani2017attention}
Vaswani, A., Shazeer, N., Parmar, N., Uszkoreit, J., Jones, L., Gomez, A.N., Kaiser, {\L}., Polosukhin, I.: Attention is all you need. Advances in neural information processing systems  \textbf{30} (2017)

\bibitem{wang2023videomae}
Wang, L., Huang, B., Zhao, Z., Tong, Z., He, Y., Wang, Y., Wang, Y., Qiao, Y.: Videomae v2: Scaling video masked autoencoders with dual masking. In: Proceedings of the IEEE/CVF Conference on Computer Vision and Pattern Recognition. pp. 14549--14560 (2023)

\bibitem{wang2017stagewise}
Wang, T., Borji, A., Zhang, L., Zhang, P., Lu, H.: A stagewise refinement model for detecting salient objects in images. In: Proceedings of the IEEE international conference on computer vision. pp. 4019--4028 (2017)

\bibitem{wang2022anchor}
Wang, Y., Zhang, X., Yang, T., Sun, J.: Anchor detr: Query design for transformer-based detector. In: Proceedings of the AAAI conference on artificial intelligence. vol.~36, pp. 2567--2575 (2022)

\bibitem{weinzaepfel2015learning}
Weinzaepfel, P., Harchaoui, Z., Schmid, C.: Learning to track for spatio-temporal action localization. In: Proceedings of the IEEE international conference on computer vision. pp. 3164--3172 (2015)

\bibitem{wu2019long}
Wu, C.Y., Feichtenhofer, C., Fan, H., He, K., Krahenbuhl, P., Girshick, R.: Long-term feature banks for detailed video understanding. In: Proceedings of the IEEE/CVF Conference on Computer Vision and Pattern Recognition. pp. 284--293 (2019)

\bibitem{wu2022memvit}
Wu, C.Y., Li, Y., Mangalam, K., Fan, H., Xiong, B., Malik, J., Feichtenhofer, C.: Memvit: Memory-augmented multiscale vision transformer for efficient long-term video recognition. In: Proceedings of the IEEE/CVF Conference on Computer Vision and Pattern Recognition. pp. 13587--13597 (2022)

\bibitem{wu2020context}
Wu, J., Kuang, Z., Wang, L., Zhang, W., Wu, G.: Context-aware rcnn: A baseline for action detection in videos. In: Computer Vision--ECCV 2020: 16th European Conference, Glasgow, UK, August 23--28, 2020, Proceedings, Part XXV 16. pp. 440--456. Springer (2020)

\bibitem{wu2023stmixer}
Wu, T., Cao, M., Gao, Z., Wu, G., Wang, L.: Stmixer: A one-stage sparse action detector. In: Proceedings of the IEEE/CVF Conference on Computer Vision and Pattern Recognition. pp. 14720--14729 (2023)

\bibitem{wu2020rethinking}
Wu, Y., Chen, Y., Yuan, L., Liu, Z., Wang, L., Li, H., Fu, Y.: Rethinking classification and localization for object detection. In: Proceedings of the IEEE/CVF conference on computer vision and pattern recognition. pp. 10186--10195 (2020)

\bibitem{xu2021long}
Xu, M., Xiong, Y., Chen, H., Li, X., Xia, W., Tu, Z., Soatto, S.: Long short-term transformer for online action detection. Advances in Neural Information Processing Systems  \textbf{34},  1086--1099 (2021)

\bibitem{zhao2022tuber}
Zhao, J., Zhang, Y., Li, X., Chen, H., Shuai, B., Xu, M., Liu, C., Kundu, K., Xiong, Y., Modolo, D., et~al.: Tuber: Tubelet transformer for video action detection. In: Proceedings of the IEEE/CVF Conference on Computer Vision and Pattern Recognition. pp. 13598--13607 (2022)

\bibitem{zhu2020deformable}
Zhu, X., Su, W., Lu, L., Li, B., Wang, X., Dai, J.: Deformable detr: Deformable transformers for end-to-end object detection. In: ICLR (2021)

\end{thebibliography}

\end{document}